\documentclass[review]{elsarticle}

\usepackage{lineno,hyperref}
\modulolinenumbers[5]

\usepackage{graphicx}
\usepackage{subcaption}
\usepackage{amsmath}
\usepackage{amssymb}
\usepackage{multirow}
\usepackage{booktabs}
\usepackage{color}

\usepackage{algorithm} %wrap algpseudocode and enrich with label etc.
\usepackage{algpseudocode} %for listing pseudocode
\usepackage{balance}
\usepackage{wrapfig}
\usepackage{lscape}
\usepackage{rotating}
\usepackage{soul}

\journal{Journal of \LaTeX\ Templates}

\begin{document}

\begin{frontmatter}

\title{Indirect and Direct Training of Spiking Neural Networks for End-to-End Control of \\ a Lane-Keeping Vehicle \footnote{\textsuperscript{\textcopyright}2020 Licensed under the Creative Commons CC-BY-NC-ND, DOI: https://doi.org/10.1016/j.neunet.2019.05.019}}

%%% Group authors per affiliation:

\author[label_1,label_2]{Zhenshan Bing}
\ead{bingzs@mail.sysu.edu.cn (cc:zhenshanhit@yahoo.com)}

\author[label_2]{Claus Meschede}
\ead{claus.meschede@tum.de}

\author[label_3]{Guang Chen}
\ead{guangchen@tongji.edu.cn
}

\author[label_2]{Alois Knoll}
\ead{knoll@in.tum.de}

\author[label_1]{Kai Huang\corref{cor1}}
\ead{huangk36@mail.sysu.edu.cn}

\cortext[cor1]{Corresponding author}
\address[label_1]{School of Data and Computer Science, Sun Yat-Sen University, China}
\address[label_2]{Department of Computer Science, Technical University of Munich, Germany}
\address[label_3]{School of Automotive Studies, Tongji University, China}

%\author{Zhenshan Bing\corref{mycorrespondingauthor}}
%\author{Claus Meschede}
%\ead{claus.meschede@tum.de}
%
%\author{Kai Huang}
%\ead{claus.meschede@tum.de}
%
%\author{Guang Chen}
%\ead{guang@in.tum.de}
%
%\author{Alois Knoll}
%\ead{knoll@in.tum.de}
%
%\author[mysecondaryaddress]{Global Customer Service}

\begin{abstract}

\textcolor{black}{
	Building spiking neural networks (SNNs) based on biological synaptic plasticities holds a promising potential for accomplishing fast and energy-efficient computing, which is beneficial to mobile robotic applications.}
However, the implementations of SNNs in robotic fields are limited due to the lack of practical training methods.
In this paper, we therefore introduce both indirect and direct end-to-end training methods of SNNs for a lane-keeping vehicle.
First, we adopt a policy learned using the \textcolor{black}{Deep Q-Learning} (DQN) algorithm and then subsequently transfer it to an SNN using supervised learning.
Second, we adopt the reward-modulated spike-timing-dependent plasticity (R-STDP) for training SNNs directly, since it combines the advantages of both reinforcement learning and the well-known spike-timing-dependent plasticity (STDP).
We examine the proposed approaches in three scenarios in which a robot is controlled to keep within lane markings by using an event-based neuromorphic vision sensor.
%All approaches have been shown to successfully follow the desired path in different scenarios with varying complexity. 
We further demonstrate the advantages of the R-STDP approach in terms of the lateral localization accuracy and training time steps by comparing them with other three algorithms presented in this paper.

\end{abstract}

\begin{keyword}
\texttt{Spiking Neural Network}\sep End-to-end Learning \sep R-STDP \sep Lane Keeping
\end{keyword}

\end{frontmatter}

\section{Introduction}
\label{sec:intro}
Utilizing robots to carry out complicated tasks with autonomy has been a realistic prospect for the future, e.g. in the fields of unmanned vehicles, social humanoid robots, and industrial inspection.
In order to acquire this advanced intelligence and operate in the real-life scenes, robots have to be able to sense their environment with sensors, which usually produce high-dimensional or large-scale data. 
Nowadays, inspired by the biological nervous system
deep learning architectures have become a promising solution, due to their superiorities for processing multi-dimensional non-liner information from training data.
\textcolor{black}{
	Yet, they differ a lot from the brain-like intelligence in both of the structural and functional properties, which make them incompatible with neuroscience findings.
	Meanwhile, due to their nature of deep architecture and substantial data, training and operating them is energy-intensive, time-consuming, and latency-sensitive.
}
%However, 
%the high computational demands of deep networks still take a toll, since training them is time consuming, energy-intensive, and typically produces high response latencies. 
Taking self-driving cars as an example, the overall computation consumes a few thousand watts, as compared to the human brain, which only needs around 20 watts of power \cite{drubach2000brain}. 
These are considerable disadvantages, especially in mobile applications where real-time responses are important and energy supply is limited.
%\textcolor{red}{
%Deep learning models are vaguely inspired by information processing and communication patterns in biological nervous systems yet have various differences from the structural and functional properties of biological brains (especially human brains), which make them incompatible with neuroscience evidences.}

%Meanwhile, deep network architectures today become a promising solution since their superiorities for extracting highly non-linear functions from training data, where classical algorithmic approaches often fail to handle the complexity involved in these problems. 
%
%Despite the hardware upgrades that make large neural networks applicable to real world problems, the high computational demands of these networks still take a toll. 
%First, training artificial neural networks is time consuming and can easily spend up to days for state-of-the-art architectures. 
%Executing large-scale trained networks is also computationally expensive and typically produces high response latencies. 
%Second, performing computations with large-scale networks on traditional hardware usually consumes a lot of energy as well. 
%In self-driving cars for example, this results in computational hardware configurations consuming a few thousand Watts compared to the human brain, that only needs around 20 Watts of Power \cite{drubach2000brain}. 
%Especially in mobile applications where real-time responses are important and energy supply is limited, these are considerable disadvantages.

A possible solution to some of these problems could be provided by event-based neural networks or spiking neural networks (SNNs) that mimic the underlying mechanisms of the brain much more realistically~\cite{kasabov2018time, bing2018survey}. 
In nature, information is processed using impulses or spikes, making seemingly simple organisms able to perceive and act in the real world exceptionally well and outperform state-of-the-art robots in almost every aspect of life~\cite{10.3389/fnsys.2015.00151}. 
SNNs are able to transmit and receive large volumes of data encoded by the relative timing of only a few spikes, which leads to the possibility of very fast and efficient computing, both in terms of accuracy and speed.
For example, human brains can perform visual pattern analysis and classification in just $100~ms$, despite the fact that it involves a minimum of 10 synaptic stages from the retina to the temporal lobe~\cite{thorpe2001spike}.
%Therefore, SNNs have tremendous potential to process information more efficiently both in terms of accuracy and speed.
%are able to transmit and receive large volumes of data encoded by the relative timing of only a few spikes, this leads to the possibility of very fast and efficient computing, both in terms of energy and data, e.g. using neurally inspired hardware such as the SpiNNaker board \citep{furber2014spinnaker} or Dynamic Vision Sensors (DVS) \citep{lichtsteiner2008128}.

%Even computations within conventional artificial neural networks (ANNs) can be understood as approximation of the impulse rate over time, event-based networks or Spiking Neural Networks (SNNs) perform computations based on individual spikes taking the precise timing into account as well.
%This approach could help processing information more efficiently both in terms of energy and data, e.g. using neurally inspired hardware such as the SpiNNaker board \cite{furber2014spinnaker} or Dynamic Vision Sensors (DVS) \cite{lichtsteiner2008128}.
%Moreover, in contrast to classical reinforcement learning algorithms, information can be processed in continuous time facilitating low latency responses without the need for time-steps \cite{fremaux2013reinforcement}.
\begin{figure}[t]
	\centering
	\includegraphics[width=0.88\textwidth]{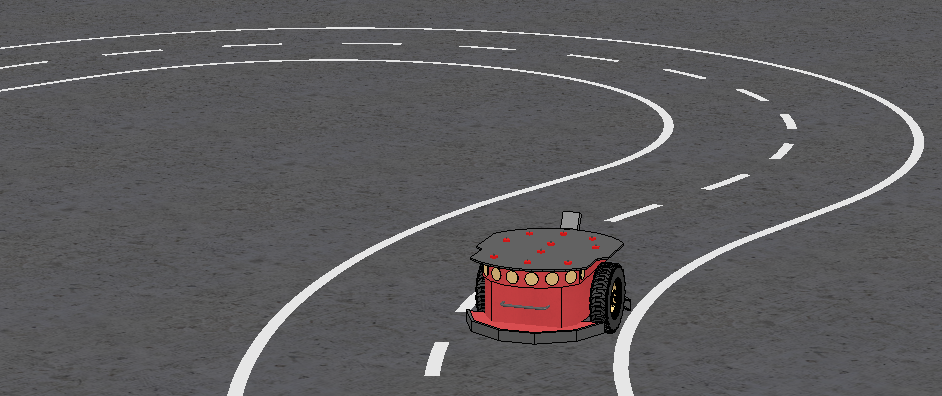}
	\caption{Robot task: lane keeping.}
	\label{fig:task}
\end{figure}

On the other hand, training these kinds of networks is notoriously difficult. 
The error back-propagation mechanisms commonly used in conventional neural networks cannot be directly transferred to SNNs due to the non-differentiabilities at spike times. 
Therefore, there has been a dearth of practical learning rules to train SNNs~\cite{Lee2016}.
Initially, SNN-based control tasks were performed by manually setting network weights, e.g. in \cite{indiveri1999neuromorphic}, \cite{lewis2000toward}, and \cite{ambrosano2016retina}.
Although this approach is able to solve simple behavioral tasks, such as wall following~\cite{wang2009wall} or lane keeping~\cite{kaiser2016towards}, it is only feasible for lightweight networks with few connections.
On the level of single synapses, experiments have shown that the precise timing of pre- and post-synaptic spikes seems to play a crucial part in the change of synaptic efficacy \cite{song2000competitive}. 
With this spike-timing-dependent plasticity (STDP) learning rule, networks have been trained in various tasks. 
For example, Wang constructed a single-layer SNN using proximity sensor data
as conditioned stimulus input and then trained it in tasks such as obstacle avoidance and target reaching~\cite{wang2008behavior, wang2014mobile}.
However, it is still not clear how the brain assigns credit as efficiently as back-propagation does, even some preliminary research has tried to bridge the gap by combining back-propagation with SNNs~\textcolor{blue}{\mbox{\cite{bengio2017stdp, Lee2016, whittington2017approximation, BOGACZ2017198}}}.

Furthermore, some research has attempted to implement biologically plausible reinforcement learning algorithms based on experimental findings in SNNs. 
Reward-modulated spike-timing-dependent plasticity (R-STDP)~\cite{Legenstein2008}\cite{florian2007reinforcement}\textcolor{blue}{\mbox{\cite{legenstein2008learning}}}, which is a learning rule that incorporates a global reward signal in combination with STDP, has recently been a research focus.
This approach intends to mimic the functionalities of those neuromodulators which are chemicals emitted in human brain, e.g. dopamine.
Therefore, R-STDP can be very useful for robot control, because it might simplify the requirements of an external training signal and leads to more complex tasks. 

However, practical robotic implementations based on R-STDP are rarely found due to its complexity in feeding sensor data into SNNs, constructing and assigning the reward to neurons, and training the SNNs.
Specifically, typical sensor data is time-based, such as data from proximity sensor and conventional vision sensor, rather than event or spike-based.
In order to feed the data into an SNN, it has to be converted into spikes somehow.
In addition, the reward should be carefully assigned to the SNN, a value that is either too high or too low will make the learning instable.
The network weights are critical for learning as well, otherwise the learning process will consume more time or even cause failures.

On the basis of our previous work~\cite{8460482}, this paper aims to explore the training algorithms for spiking neural networks from two different ends and implement them for end-to-end control in the robotics domain.
\textcolor{black}{
	We conduct our research in four parts.}
First, we construct a simulated lane scenario and adapt it with different lane patterns for evaluating different algorithms, in which a pioneer robot mounted with a dynamic vision sensor (DVS)~\cite{lichtsteiner2008128} is deployed to perform the task.
\textcolor{black}{
	The DVS directly outputs event-based spikes when there is a change of illumination on the pixel level. 
	Thus, it fits SNNs well due to its spike-based nature and offers some great advantages over traditional vision senors, such as speed, dynamic range, and energy efficiency~\cite{lichtsteiner2008128}.}
Second, in an indirect training setup, a conventional ANN is trained in a classic reinforcement learning setting using the Deep Q-Learning (DQN) algorithm. 
Afterwards, the learned policy is transferred to train an SNN on \textcolor{black}{a state-action dataset created by collecting data from the RL scenarios} using supervised learning.
Third, an event-based neural network is constructed using the STDP dopamine synapse model
and directly trained by the R-STDP learning rule.
The reward given to the SNN is defined for each motor individually as a linear function of the lane center distance.
Finally, we compare the training performances of all four networks by running them in the training and testing scenarios.

\textcolor{black}{
%We propose both indirect and direct end-to-end training approaches with respect to reinforcement learning paradigm in SNNs for robotic control tasks.
Our main contributions to the literature are summarized as follows.
First, our indirect approach utilizes the learned knowledge from a classical reinforcement learning setting and successfully transfers it into an SNN-based controller.
This transition offers a way to quickly build up an applicable spike-based controller on the basis of conventional ANNs in robotics, which can be further executed on a neuromorphic hardware to achieve fast computation.
%First, our indirect approach creates a dataset with the learned knowledge from a conventional ANN trained in a classic reinforcement learning setting using the DQN algorithm.
%Then, we transfers the knowledge to an SNN-based controller by training it with the dataset using supervised learning.
%This transition offers a way to quickly build up an applicable spike-based controller on the basis of conventional ANNs in robotics, which can be further executed on a neuromorphic hardware to achieve fast computation.
Second, our direct approach trains the SNN with the R-STDP learning rule in a biologically plausible way and demonstrates fast and accurate learning process when taking the advantages of an event-based vision sensor. 
This approach resembles the neural modulation process, which serves as one of the main functionalities in brains and is responsible for strengthening the synaptic connections and then reinforcing desired behaviors or actions. 
Third, by comparing the performances of all controllers, we demonstrate the superiorities of the R-STDP approach in terms of the training time steps, lateral localization accuracy, and adaption to unknown challenging environment. Those advantages make this method very suitable for being used in mobile robots applications, which usually require quick learning ability and environmental adaptation.
}

The remainder of the paper is organized as follows: Section~\ref{sec:task} introduces the simulation setups for performing the lane-keeping tasks.
Section~\ref{sec:dqn} present the indirect learning method for transferring the policy from DQN to SNN.
Section~\ref{sec:controller} presents the implementation details of the direct training based on R-STDP. In Section~\ref{sec:discuss}, the training details of each controllers are presented.
In Section~\ref{sec:performance}, we provide the experimental results and make a comparison to other controllers.
Section~\ref{sec:conclusion} summarizes our study and presents the future work.

\section{Lane-Keeping Tasks}
\label{sec:task}

\begin{figure}[t]
	\centering
	\includegraphics[width=0.8\textwidth]{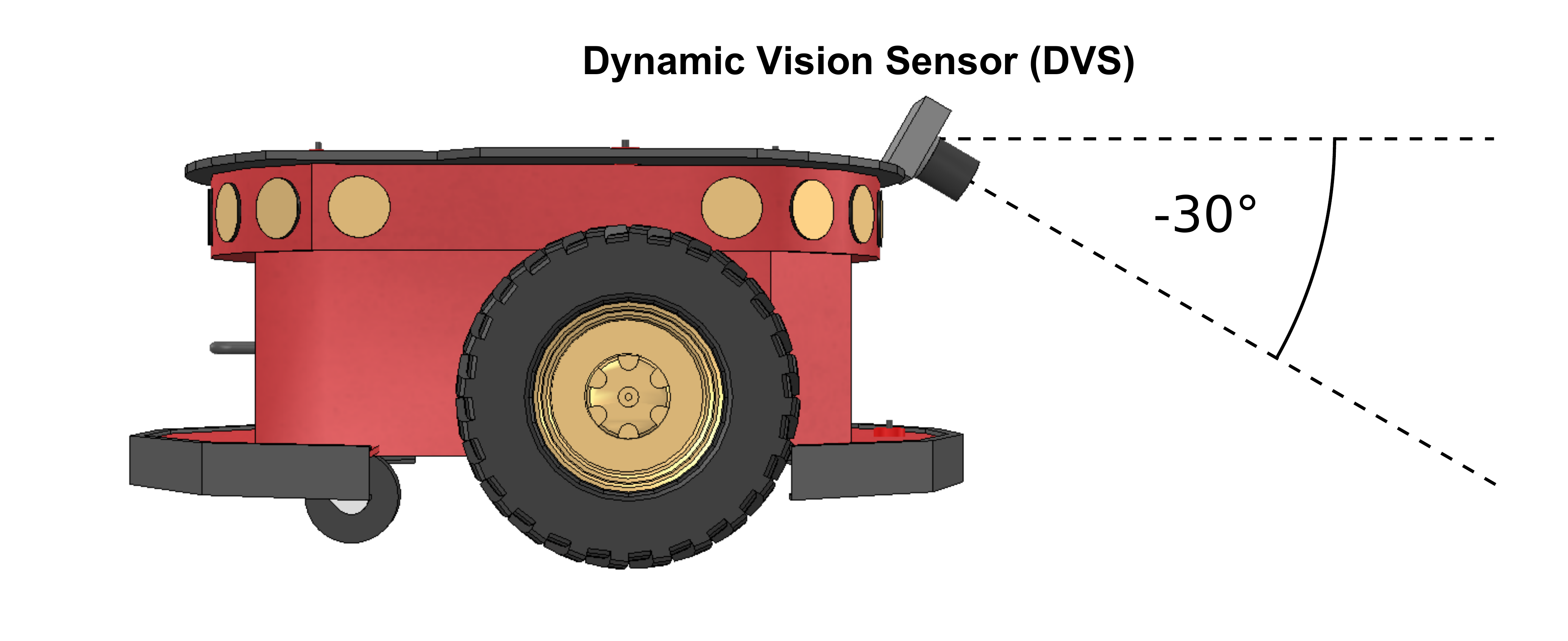}
	\caption[Pioneer P3-DX robot.]{Pioneer P3-DX robot with dynamic vision sensor (DVS).}
	\label{fig:pioneer}
\end{figure}

\begin{figure}[t]
	\centering
	\includegraphics[width=0.76\linewidth]{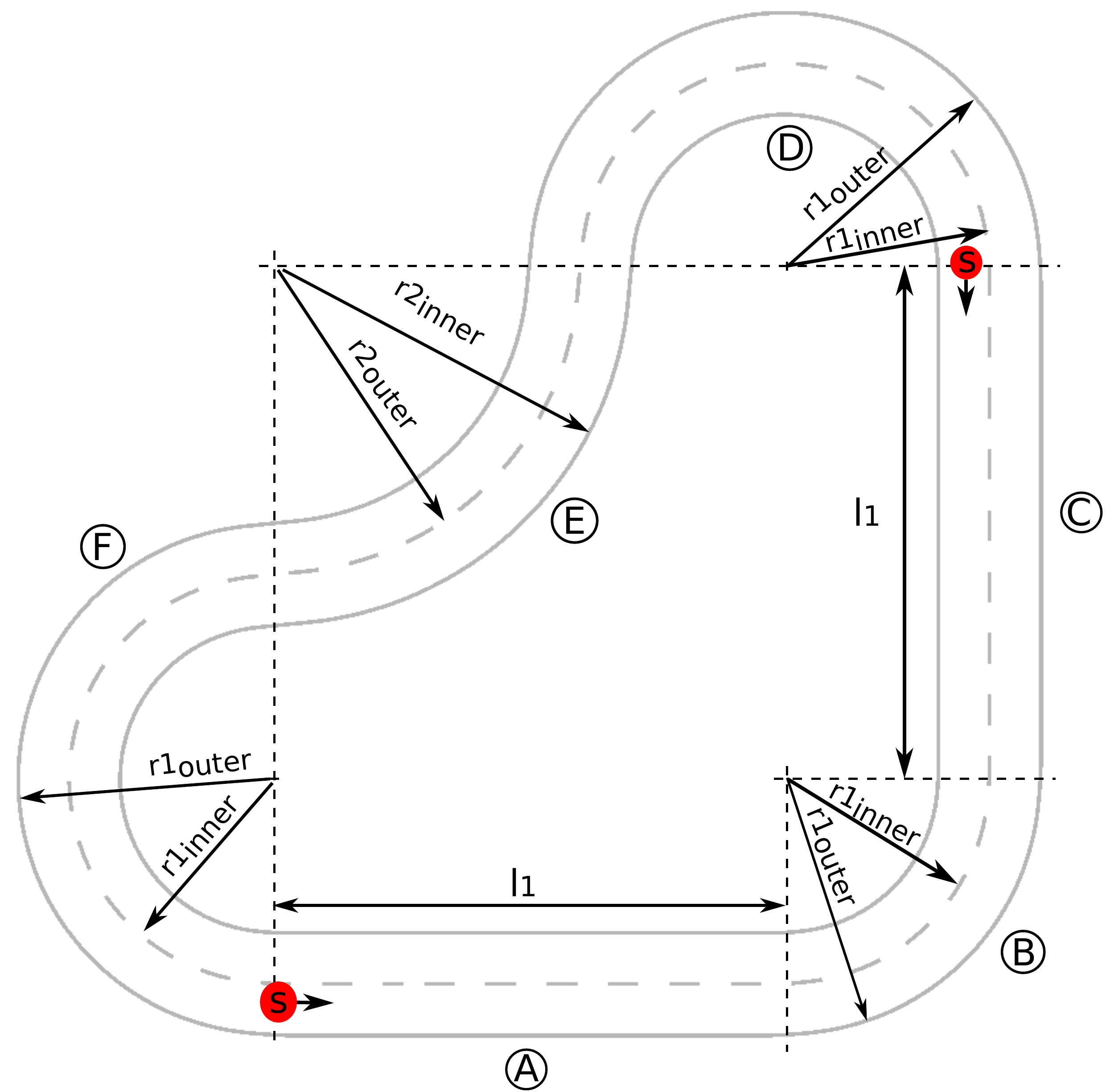}
	\caption{Scenario 1: The simple lane-keeping scenario consists of a road with two lanes (inner and outer) and six different sections A, B, C, D, E, and F. Starting positions are marked with $S$. Dimensions: $r1_{inner} = 1.75~m$, $r2_{inner} = 3.25~m$, $r1_{outer} = 2.25~m$, $r2_{outer} = 2.75~m$, $l_1 = 5.0~m$.}
	\label{fig:scenario_sub1}
\end{figure}

\begin{figure*}[t]
	\centering
	\begin{subfigure}[t]{.35\textwidth}
		\includegraphics[width=\linewidth]{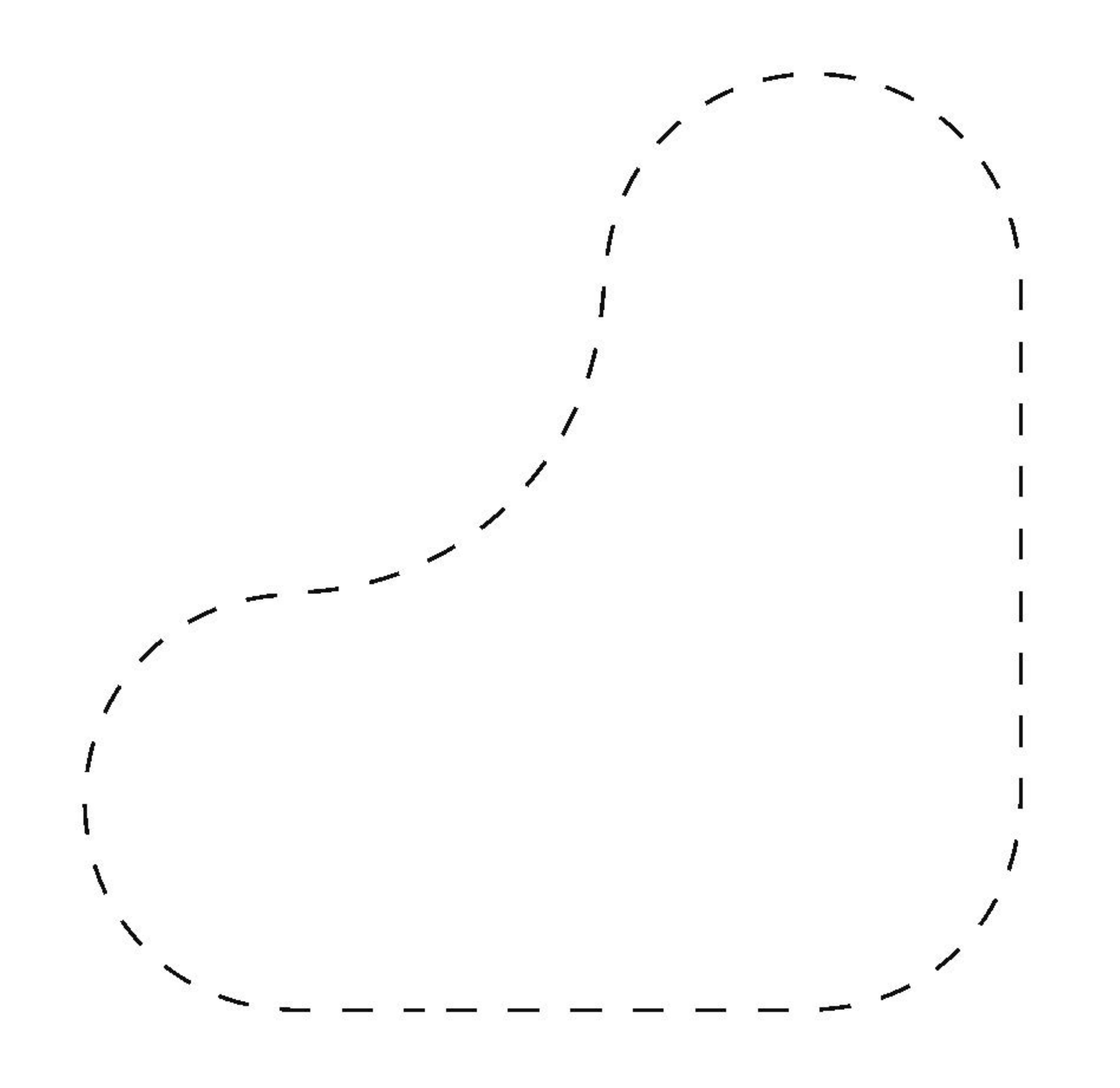}
		\caption{Scenario 2}
		\label{fig:scenario_sub2}
	\end{subfigure}
	\begin{subfigure}[t]{.35\textwidth}
		\includegraphics[width=\linewidth]{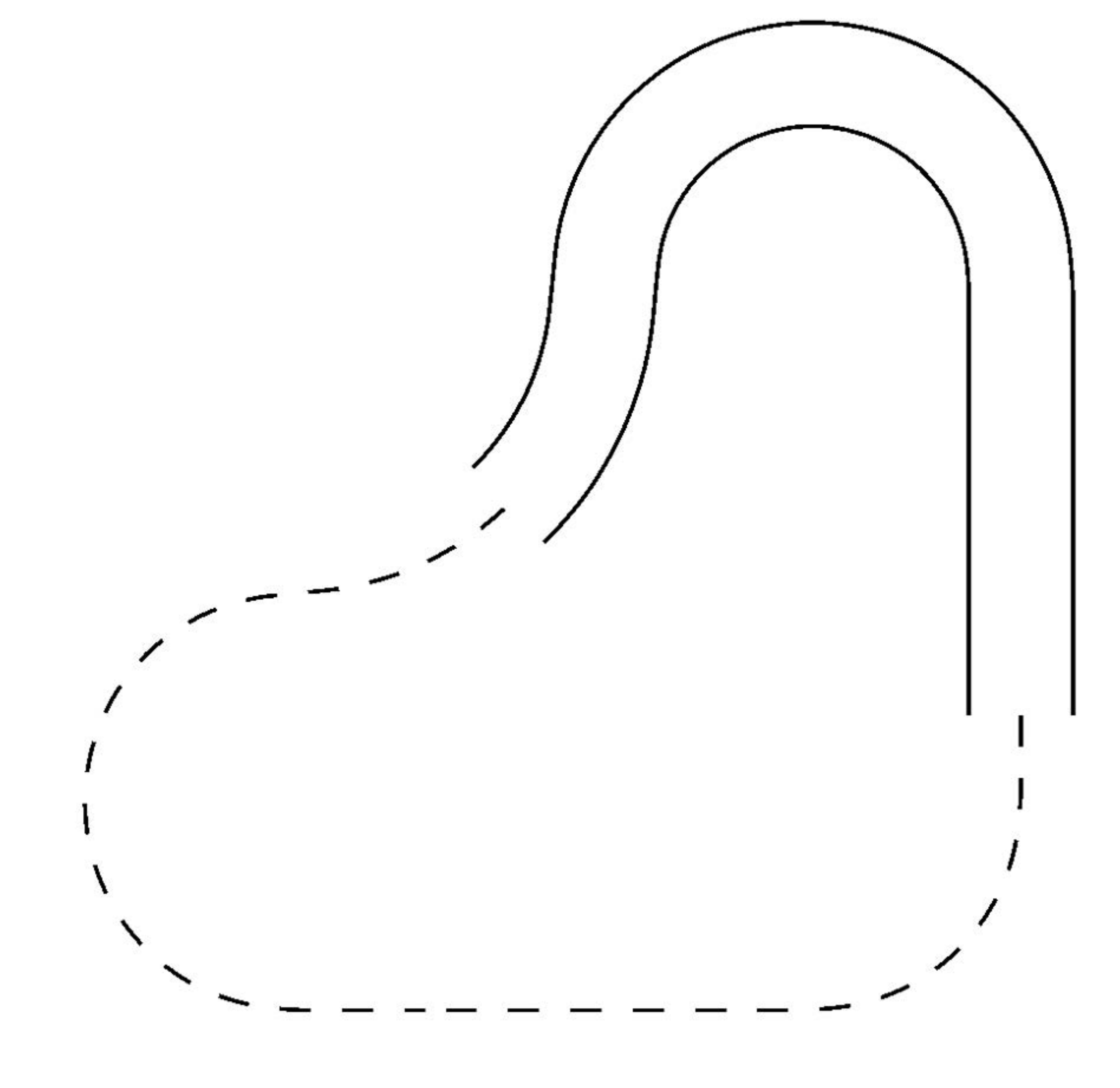}
		\caption{Scenario 3}
		\label{Lanes with two different patterns}
	\end{subfigure}
	\caption{(a) Scenario 2: Single lane pattern without boundaries. (b) Scenario 3: Lanes with two different patterns.}
	\label{fig:scenario}
\end{figure*}

In order to provide a simple and flexible environment to test and compare different algorithms, simulated lane-keeping tasks with different lane patterns for a Pioneer robot are set up using Virtual Robotics Experiment Platform (V-REP)~\cite{rohmer2013v} (See Fig.~\ref{fig:task}). All the sensor messages and motor commands between the simulator and the neural networks are transmitted via ROS~\cite{quigley2009ros}.

%\subsection{Simulation Environment}
%\label{sec sim env}

%In order to provide a simple and flexible environment to test and compare different algorithms, the cross-platform robotics simulator V-REP~\citep{rohmer2013v} is used in this paper. 
%To communicate with other computer programs outside the simulator, V-REP provides a ROS~\cite{quigley2009ros} interface plugin for creating ROS nodes for publishing and subscribing to ROS topics.

%A right lane following scenario for a Pioneer robot is shown in Fig.~\ref{fig:task}.
Instead of using the on-board ultrasonic sensors, a DVS camera is attached to the front of the robot with a $30^{\circ}$ depression angle as shown in Fig.~\ref{fig:pioneer}.
For further validating the effectiveness and adaptability of the proposed algorithms, three scenarios with different lane patterns are shown in Fig.~\ref{fig:scenario}.
The first scenario in Fig.~\ref{fig:scenario_sub1} consists of a closed loop course with a two-lane road.
The road is comprised of two solid lines and a uniformly dashed line in the middle.
From the starting position onwards, the outer lane can be divided into six sections: $(A)~straight$, $(B)~left$, $(C)~straight$, $(D)~left$, $(E)~right$, $(F)~left$.
During each episode in the training, the robot will switch the start position and
moving direction between inner and outer lane at each reset.
Therefore, it will experience both left and right turns equally and with different radii as well. 

Based on the same layout and dimensions, a second scenario was implemented, testing the algorithms on a different road pattern where the left and right solid lines are missing (See Fig.~\ref{fig:scenario_sub2}).
In a third scenario, two different road patterns have to be learned in parallel (See Fig.~\ref{Lanes with two different patterns}).

\section{Indirect Learning based on DQN}
\label{sec:dqn}

In this section, we will first solve the lane-keeping tasks using a classic Deep Q-Learning (DQN) reinforcement learning algorithm.
Then we will introduce the indirect training method by transferring the learned policy from the DQN to an SNN within the framework of supervised learning.
All the simulation parameters can be found in the tables in the appendix.

\begin{figure}[h]
	\centering
	\includegraphics[width=0.8\textwidth]{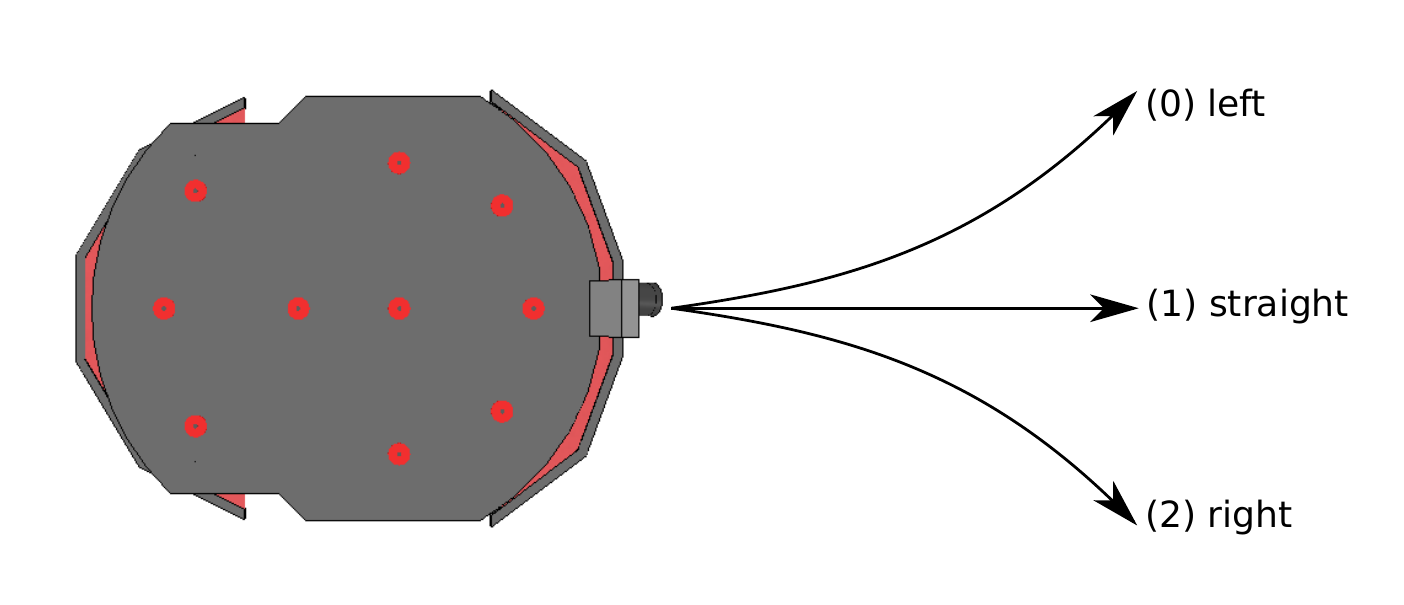}
	\caption{Action space in lane-keeping task with three discrete actions: (0) Turn left: Set left motor
		speed to $v_s-v_t$ and right motor speed to $v_s + v_t$. (1) Go straight: Set left and right motor speed to $v_s$. (2) Turn right: Set left motor speed to $v_s+v_t$ and right motor speed to $v_s-v_t$.}
	\label{fig:action}
\end{figure}

\subsection{Lane Keeping as MDP}
Reinforcement learning tasks are usually described as a Markov decision process (MDP), which is defined as a 5-tuple of actions, states, transition probabilities, rewards, and discount factor.
While the transition probabilities can be ignored when using model-free algorithms such as Q-learning, other components of the MDP have to be carefully chosen to ensure fast and stable learning.

Fig.~\ref{fig:action} shows three discrete actions that the robot can take for these tasks. 
It can go straight, letting left and right motors run at the same speed, or it can take a turn by adding or subtracting speed to both motors depending on the desired moving direction.

\subsection{DVS Input Generation}
\label{sec dvs sim}
\begin{figure}[t]
	\centering
	\includegraphics[width=0.88\textwidth]{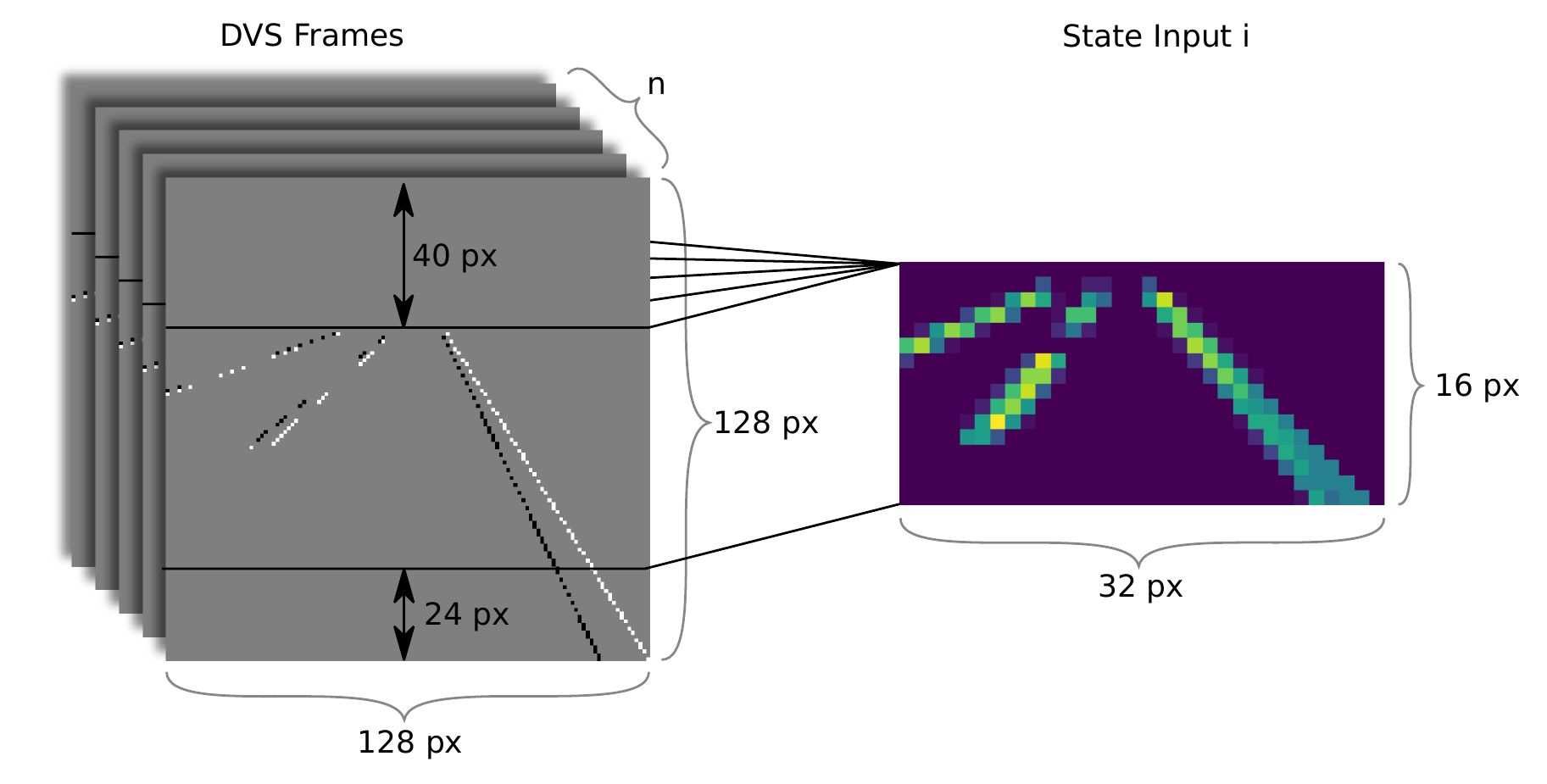}
	\caption[State space.]{Conversion of consecutive DVS frames into state input for reinforcement learning. This is done by dividing the original $128 \times 128$ DVS frames into small $4 \times 4$ regions and counting every event over consecutive frames regardless of the polarity. Furthermore, the image is cropped at the top and bottom, resulting in a $32 \times 16$ image.}
	\label{fig:image_preproess}
\end{figure}

%\textcolor{red}{
%In order to reduce noise in the images, the DVS camera can only detect the road markings and other deliberately placed objects in the simulation. 
%Intensity changes on the ground for example are ignored.}
In similar reinforcement learning tasks using conventional cameras~\cite{mnih2015human,lillicrap2015continuous}, scaled images could be directly used as state input for the MDP; this is more difficult when using a DVS device.
Dynamic Vision Sensor, as an emerging neuromorphic sensor, generates sparse, event-based output that represents the positive and negative relative luminance change of a scene.
Due to its advantages, such as speed, dynamic range, and energy efficiency~\cite{lichtsteiner2008128}, DVS is used in this study to detect the lane marks and generate spikes.
First, in order to decrease the computational complexity of the task, images are reduced to a lower resolution as well. 
Second, due to the event-based nature of the DVS data, image frames coming from the simulation do not always contain sufficient information for the network to make meaningful decisions. 
Therefore, the state input is computed by condensing information from several consecutive DVS frames into a single image. 
\textcolor{black}{To be clear, the DVS frame means the events accumulated in the $50~ms$ interval instead of being the whole image frame as traditional vision sensor.}
As shown in Fig.~\ref{fig:image_preproess}, this is done by dividing the original $128 \times 128$ DVS frames into small $4 \times 4$ regions and counting every event over consecutive frames regardless of the polarity. 
Furthermore, the image is cropped at the top and bottom, resulting in a $32 \times 16$ image. 
To further increase the performance, the final DQN state input $s_{M \times N}$ is a binary version of the state input $i_{M \times N}$, only containing ones and zeros:
\begin{equation}
	\centering
	s_{M\times N}=\begin{cases}
	0 & \text{ if } i_{M\times N}= 0\\ 
	1 & \text{ if } i_{M\times N}>0 
	\end{cases}
\end{equation}

In the simulation, DVS frames are calculated and published every 50 \textit{ms} (with every simulation time step). 
Actions are executed every 500 \textit{ms}. 
Therefore, during one action step, DVS frames are stored in a first-in first-out (FIFO) queue of length 10, and the last ten DVS frames are then converted into the final state input.

\subsection{Reward Generation for DQN}

\begin{figure}[t]
	\centering
	\includegraphics[width=0.88\textwidth]{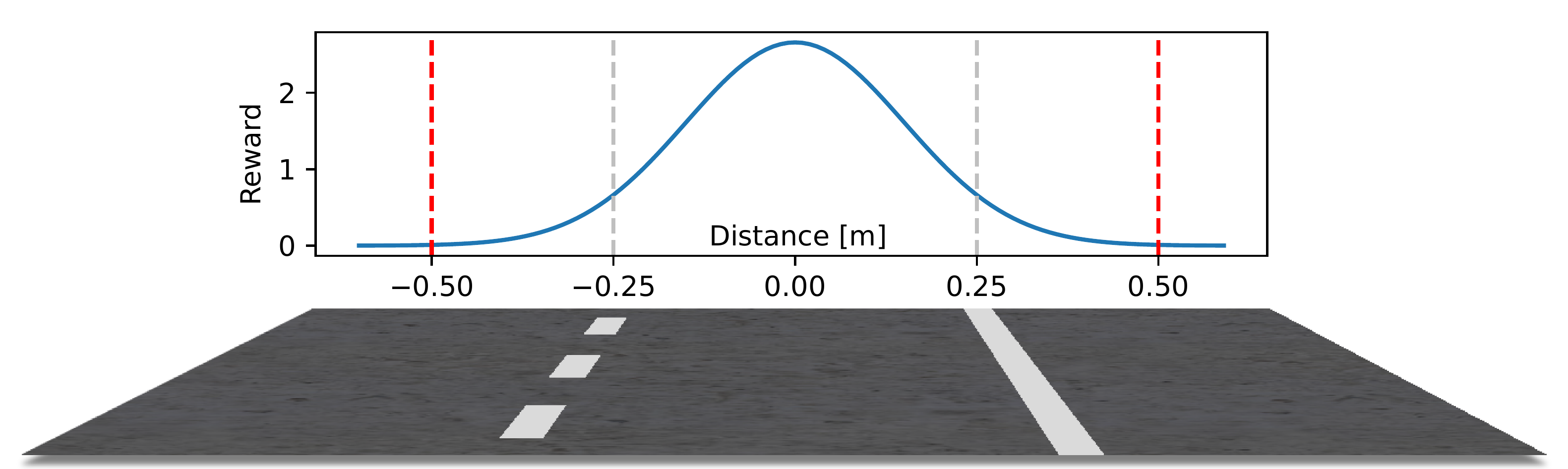}
	\caption[Reward.]{Reward given in the lane-keeping task. It is defined as a Gaussian distribution over the distance to the center of the lane with a standard deviation of $\sigma = 0.15$ and mean at $0$. The lane markings are $0.25\,m$ away from the lane center. If the robot goes further than $0.5\,m$ from the lane center, episodes are terminated and the robot is positioned at its starting position.}
	\label{fig dqn reward}
\end{figure}

Rewards play a crucial role in reinforcement learning and define the goal of an agent. In this research study, the robot is supposed to learn to follow a lane staying as close to the center as possible. 
Fig. \ref{fig dqn reward} shows the definition of the reward that is given at every time step of the MDP. 
It is defined as a Gaussian distributed function of the lane center distance. 
As the model-free DQN algorithm learns from experience samples with a one-step lookahead, it is beneficial for learning to use a reward that is well distributed over the state space and monotonically increasing towards the goal. 
This ensures that the robot will learn to navigate in the direction of the goal, even if it has not been there yet. Besides DVS data, the simulator publishes position data of the robot every $50\,ms$ as well. 
With a mathematical model of the lane center in both directions, this data is used to calculate the exact distance of the robot to the lane center and the resulting reward. 

If the robot reaches a position where its distance to the lane center is greater than $0.5\,m$, training episodes are terminated and a reset message is generated that causes the simulator to place the robot at its starting position on the opposite lane. Letting the robot alternate between both lane directions increases its experienced states and results in a more generalized policy after learning. 
In reinforcement learning, the extent to which an agent takes expected future rewards into account is usually controlled by a discount factor. 
Although the lane-keeping task does not necessarily require looking ahead many steps, the discount factor is set to $0.99$, therefore potentially being able to solve tasks that involve some foresight as well.

\subsection{DQN-based Controller}

A fully connected \textcolor{black}{feedforward} network architecture using rectified linear units (ReLU) \textcolor{black}{as activation function was chosen, inspired by} similar work~\cite{lee2016training,7280696}.
The network takes the binary state image as input, resulting in $32 \times 16 = 512$ input neurons. It consists of two hidden layers with 200 neurons each and three output neurons representing the discrete actions. 
Training is performed using the stochastic optimization algorithm Adam.

%\begin{algorithm}
%	\caption{Algorithm for DQN}
%	\begin{algorithmic}[1]
%		\renewcommand{\algorithmicrequire}{\textbf{Input:}}
%		\renewcommand{\algorithmicensure}{\textbf{Output:}}
%		\REQUIRE in
%		\ENSURE  out
%		\\ \textit{Initialisation} :
%		\STATE first statement
%		\\ \textit{LOOP Process}
%		\FOR {$i = l-2$ to $0$}
%		\STATE statements..
%		\IF {($i \ne 0$)}
%		\STATE statement..
%		\ENDIF
%		\ENDFOR
%		\RETURN $P$ 
%	\end{algorithmic} 
%\end{algorithm}
%
%\begin{figure*}[t]
%	\centering
%	\hspace*{-0.2cm}
%	\includegraphics[width=0.7\textwidth]{figures/dqn.pdf}
%	\caption[DQN flowchart.]{Flowchart of the DQN algorithm.}
%	\label{fig dqn flow}
%\end{figure*}

\begin{figure}[!tp]
	\centering
	%\hspace*{-0.2cm}
	\includegraphics[width=0.69\textwidth]{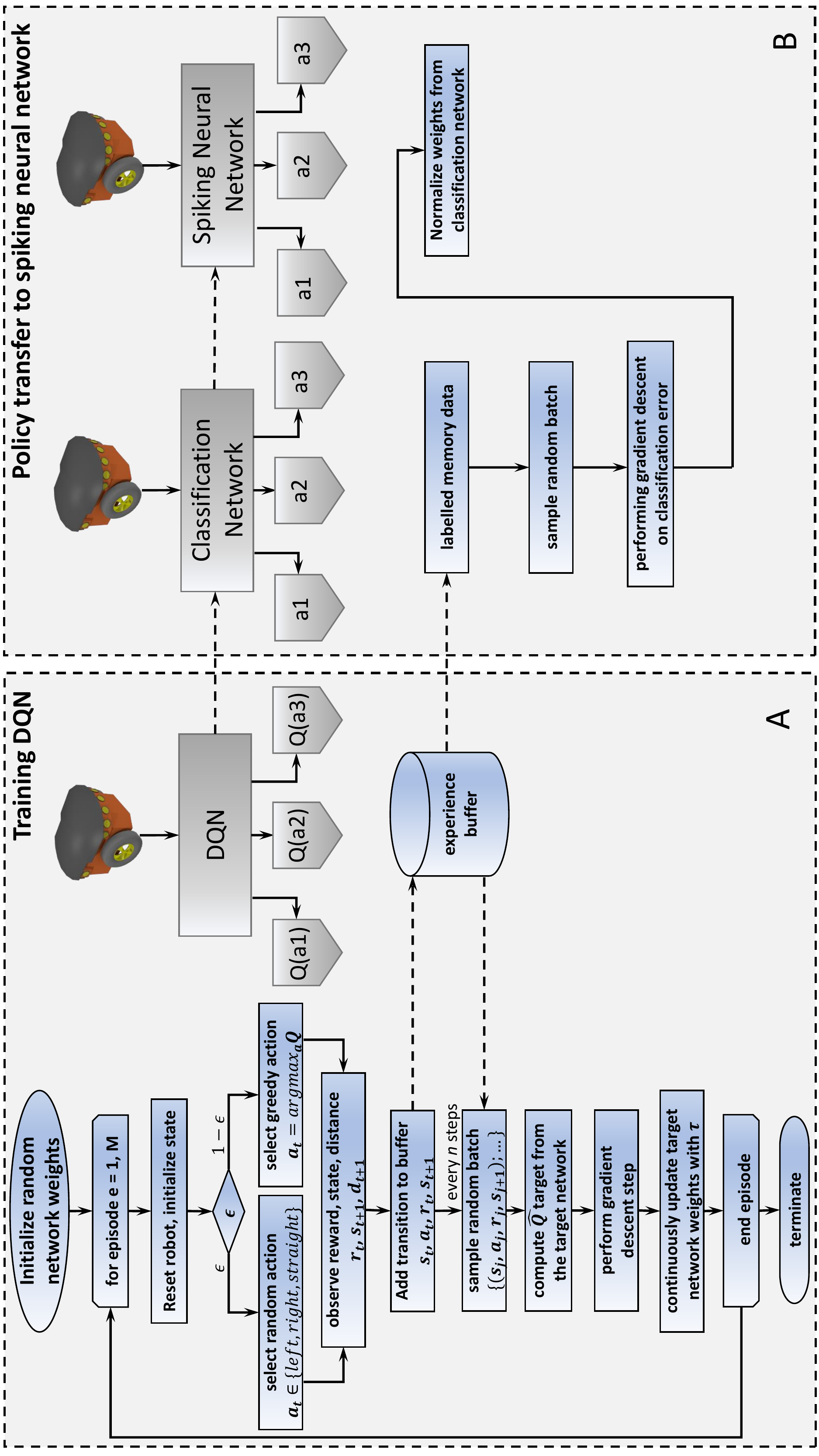}
	\caption[DQN flowchart.]{A. Flowchart of the DQN algorithm. B. Flowchart of the policy transfer from DQN to SNN.}
	\label{fig flow chart}
\end{figure}

%\begin{sidewaysfigure}[ht]
%	\includegraphics[width=0.5\textwidth]{../IEEEtran/figures/flow_chart.pdf}
%	\caption[DQN flowchart.]{A. Flowchart of the DQN algorithm. B. Flowchart of the policy transfer from DQN to SNN.}
%	\label{fig flow chart}
%\end{sidewaysfigure}

Fig.~\ref{fig flow chart}A shows a detailed flow chart of the DQN algorithm. 
\textcolor{black}{There are two networks being used in the DQN algorithm, which share the same architecture but with different weight parameters.
The action-value network $\mathcal{Q}$ is used to determine the action with the highest $Q$ value, of which the weights are updated every step.
The target action-value network $\hat{\mathcal{Q}}$ is used to predict the $q\_target$ value, of which the weights will be updated every several steps by assigning the weights as $\hat{\mathcal{Q}} = \mathcal{Q}$.}
More details about the DQN algorithm itself can be found in~\cite{van2016deep} 
At the beginning, the \textcolor{black}{action-value} network $\mathcal{Q}$ is initialized with random weights and copied to the \textcolor{black}{target action-value} network $\hat{\mathcal{Q}}$. 
Each episode of the training procedure begins with a reset of the robot to its starting position, switching lanes after each episode. 
Hereby, the initial state input is a vector of zeros.
At each time step, actions are chosen from $\mathcal{Q}$ following an $\epsilon$-greedy policy. 
This means that with a probability of $\epsilon \in [0, 1]$ the agent will randomly select an action. 
Otherwise, it will select the action with the highest action value. 
At the start, $\epsilon$ is set to 1, ensuring pure exploratory behavior. 
After a predefined $1{,}000$ time steps, $\epsilon$ then linearly decreased to its end value close to zero.

Chosen actions \textcolor{black}{$a_t$} are sent to the environment handler, which will communicate with the simulator to acquire the reward \textcolor{black}{$r_t$}, next state image \textcolor{black}{$s_{t+1}$}, and the distance to the lane center \textcolor{black}{$d_t$}. 
Moreover, each transition (\textcolor{black}{$s_t$, $a_t$, $r_t$, $s_{t+1}$}) is stored in the experience buffer. Every $n$ steps, the actual training step is performed by randomly sampling transitions from the experience buffer. 
Using the target action-value network $\hat{\mathcal{Q}}$ for calculating the updating targets, the loss function is then constructed in order to perform a stochastic gradient descent step on the action network. 
At the end of each training step, the target action-value network $\hat{\mathcal{Q}}$ weights are gradually updated towards
the action-value network $\mathcal{Q}$ weights with $\tau \in [0, 1]$  and $\tau \leqslant 1$ ~\cite{lillicrap2015continuous}.

Training episodes will be terminated if the robot goes beyond the maximum distance to the lane center or if the maximum number of steps in an episode is reached. 
The latter mechanism guarantees that the robot will experience both directions of the road, even if it has already learned a good policy for keeping the lane. 
The overall training procedure is ended either after reaching a predefined number of episodes or total training steps.

\subsection{DQN-SNN based Controller}
The aforementioned DQN algorithm handles event-based data by storing consecutive DVS frames and batch processing them at every step in the MDP. 
Clearly, this approach cannot be the ideal mechanism for handling DVS data, as it annihilates some of the advantages that make the sensor powerful in the first place, e.g. its temporal resolution. 
However, handling data streams is precisely what the SNNs are good at, e.g. from a DVS device, without the need for batch processing.
%Spiking Neural Networks could help solve these problems. 
%First, time plays as crucial role in these networks, which makes them suitable for handling data streams, e.g. from a DVS device, without the need for batch processing. 
%Second, their neurally inspired design could lead to efficient hardware for power efficient computing like SpiNNaker.
Due to their event-based nature, spikes have to be decoded somehow in order to obtain real values, which makes it very difficult to get network output with similar precision as well as DQN. 
Moreover, the non-differentiability of spike events makes it very difficult to use a training mechanism such as back-propagation.
%, even though Lee et al. [10] have recently shown how this could be achieved anyhow. 
%In classification tasks on the other hand where precise output spike decoding is not necessary, SNNs have repeatedly shown to generate good results even when compared to conventional ANNs~\cite{}.
Therefore, in this sub-section we show how the previously learned DQN policy can be used to create \textcolor{black}{a state-action dataset created by collecting from the RL scenario} for training an SNN using supervised learning (Fig.~\ref{fig flow chart}B). 

To address this problem, ReLU is considered as an activation function of the input stimulus and the firing frequency, rather than the function of the input stimulus and the action potential, since there is a linear relationship between the action potential and the firing frequency.
Based on the fact that simple integrate-and-fire (IF) neurons in SNNs behave very similarly to rectified linear units (ReLU) in conventional ANNs, an indirect training method can be used for training:
\begin{enumerate}
	\item Create the state-action dataset by labeling stored states with corresponding actions.
	\item Train conventional ANN with no hidden layer biases and ReLU activation functions.
	\item Normalize weights.
	\item Transfer weights to SNN with IF neurons and perform control task.
\end{enumerate}

For training ANNs using stochastic gradient descent on the prediction error, the DQN algorithm stores every single transition in the previously discussed experience buffer. This makes it very convenient to use the same data for training the SNN as well. 
Therefore, first, all stored state images as shown in Fig.~\ref{fig:image_preproess} are labeled using the pre-trained action network from DQN. 
It is important to note that the input images $s_{M \times N}$ for previously training the DQN are still used here to train the SNN.
Furthermore, the pixel values $i_{m,n}$, describing the number of spike events in the same $4 \times 4$ window over consecutive DVS frames, are scaled to
$\hat{i}_{m,n} \in [0, 1]$ by dividing every value by the maximum pixel value $i_{max} = max_{j,m,n} (i^{j}_{m,n})$ in the whole
dataset. 
As a result, the input values can be interpreted as spike firing rates making the network transferable to an SNN.

In the next step, the labeled dataset is used to train a conventional ANN. 
The \textcolor{black}{fully connected feedforward} network consists of an input layer with bias and $32\times 16 = 512$ input neurons, a hidden layer with $200$ neurons, and an output layer with three output neurons.
%All the neurons in the hidden layer and the output layer are without bias.
\textcolor{black}{
	For converting an ANN to an SNN, we would like all the neurons in the hidden layer and the output layer are only effected by the activities of the neurons in the previous layer.
	Then, we can scale the weights only according to the threshold value, which is used for all the neurons in the same layer and do not have to worry about the bias value for each individual neuron.
	Therefore, all the neurons in the hidden layer and the output layer are set without bias.}
After training, the weights can be transferred to an SNN with IF neurons that matches the previous network architecture.
There is a possibility that the inputs will stimulate the hidden neurons, firing immediately in a single simulation time step.
In this case, the information from inputs cannot be precisely transmitted and indicated by the firing rate of the hidden neurons, which may cause information loss for the output layer.  
Therefore, we have to make sure that all the neurons can only fire once in each time step and then ensure minimal accuracy reductions in the SNN with transferred weights. 
This can be done by scaling weights so that the maximal weighted input to a neuron in each layer is equal	 to the firing threshold.
In other words, the weights are normalized for each layer separately beforehand.

The normalization algorithm is explained in~\cite{7280696} and shown in Algorithm~\ref{fig snn alg}. 
For the first layer, the bias is necessary to handle input vectors consisting of zeros only.
Without any bias, the network output would be always zero as well, regardless of its weights. 
In multi-layered SNNs, external input currents introducing biases to deeper layers are difficult to handle, because they have to be scaled to match the firing rates coming from previous layer neurons. 
For the first layer though, the bias can be interpreted as an additional input current with a constant firing rate of $1$.

The SNN with simple IF neurons is implemented, inspired by \cite{7280696}, which uses a time-step-based approach in order to propagate spikes through the network and update membrane potentials. 
As mentioned earlier, the simulator publishes DVS data every $50\,ms$. The data is then scaled to \textcolor{black}{[0,~1]}, causing Poisson input neurons \cite{stevens1996integrate} to fire for $50\,ms$ as well. 
The scaling factor can be roughly estimated by dividing the maximum pixel value $v_{max}$ by the number $n$ of consecutive DVS frames used in the data set. 
Therefore, if a Poisson neuron fires with its maximum frequency over $n$ simulation steps, it can be interpreted as the maximum firing rate of $1$ in the data set.

\begin{algorithm}[t]
	\begin{algorithmic}[1]
		%\For {layer}{layers:}
		\For {$layer$ in $layers$}
		\State {$max\_input = 0$}
		%\For {neuron}{layer.neurons:}
		\For {$neuron$ in $layer.neurons$}
		%\Comment{Find maximum input for this layer}
		\State {$input\_sum = 0$}
		% \For {input\_wt}{neuron.input\_wts:}
		\For {$input\_wt$ in $neuron.input\_wts$}
		\State {$input\_sum \mathrel{+}= \max (0,\; input\_wt)$}
		\EndFor
		\State {$max\_input = \max (max\_input,\;input\_sum) $}
		\EndFor
		%\For {neuron}{layer.neurons:}
		\For {$neuron$ in $layer.neurons$}
		%\Comment{Rescale all weights}
		%\For {input\_wt}{neuron.input\_wts:}
		\For {$input\_wt$ in $neuron.input\_wts$}
		\State {$input\_wt = input\_wt\; /\; max\_input$}
		\EndFor
		\EndFor
		\EndFor
	\end{algorithmic}\caption{Model-Based Normalization~\cite{diehl2015fast}}
	%\caption[Model-based normalization algorithm.]{Model-based normalization algorithm for converting ANNs into SNNs according to \cite{diehl2015fast}.}
	\label{fig snn alg}
\end{algorithm}

The information from consecutive DVS frames is propagated through the SNN over time. When a neuron fires and sends a spike to the next layer, it increases the membrane potential of the next layer's neurons. 
Therefore, information is stored in the membrane potentials and it takes some time to generate output spikes. 
As a consequence, this means that output spikes are generated sparsely in time, leaving simulation steps with no spike output at all. 
In order to generate a control signal, even if there are no output spikes during a simulation step, a trace is implemented for each action. 
The action trace $z_t^a$ accumulates output spikes $s_t$ for each action respectively and decays over time with a factor \textcolor{black}{\mbox{$c \in [0,~1]$}}. 
The action $a_t$ with the highest trace value is eventually chosen at every simulation step:

\begin{align}
z_{t+1}^a = c\cdot z_t^a + s_t\\
a_t = \text{argmax}_a (z_t^a)
\label{equ:trace}
\end{align}

%Figure \ref{fig comm snn} shows all components that have been used for the SNN controller. The communication with the simulator is managed using the same environment handler as before.

\section{Direct Learning of SNN with R-STDP}
\label{sec:controller}

Training a neural network with DQN to learn a policy and transferring the policy to a SNN by creating an labeled state-action dataset is cumbersome, and it introduces some loss in the training process. 
Furthermore, this approach ignores one of the main strengths that SNNs bring compared to conventional ANNs, which is their ability to take the precise timing of spikes into account and not just the averaged rate.	
To tackle this problem, an SNN is constructed and trained using R-STDP for steering the robot in the aforementioned lane-keeping tasks.

\subsection{R-STDP Learning Rule}
As the most important theory in neuroscience explaining the adaption of synaptic efficacies in the brain during the learning process, the spike-timing-dependent plasticity (STDP) learning rule~\cite{caporale2008spike} has been successfully proven by neuroscience experiments~\cite{markram1997regulation, bi1998synaptic}.
% and applied to problems such as input clustering, pattern recognition, and self-organizing maps~\cite{hinton1999unsupervised}.
%This phenomenon has been termed Spike-Timing-Dependent-Plasticity (STDP) to expound the activity-dependent development of nervous systems.

For this study, the weight update rule under STDP as a function of the time difference between pre- and postsynaptic spikes is defined as
\begin{equation}
\centering
\Delta t = t_{post} - t_{pre} 
\end{equation}
\begin{equation}
\centering
W(\Delta t) =\begin{cases}A_+ e^{-\Delta t / \tau_+}, \text{ if } \Delta t \geq 0 \\
- A_- e^{\Delta t / \tau_-} , \text{ if } \Delta t < 0 \end{cases} 
\label{eq_stdp}
\end{equation}
\begin{equation}
\centering 
\Delta w = \sum_{t_{pre}} \sum_{t_{post}} W(\Delta t)
\end{equation}
, where $w$ is the synaptic weight. $\Delta w$ is the change of the synaptic weight.
$t_{pre}$ and $t_{post}$ stand for the timing of the firing spike from pre-neuron and post-neuron.
$A_+$ and $A_-$ represent positive constants scaling the strength of potentiation and depression, respectively. $\tau_+$ and $\tau_-$ are positive time constants defining the width of the positive and negative learning window. 

A simple learning rule combining models of STDP and a global reward signal was proposed by Izhikevich~\cite{izhikevich2007solving} and Florian~\cite{florian2007reinforcement}.
In the R-STDP, the synaptic weight $w$ changes with the reward signal $R$.
The eligibility trace of a synapse can be defined as,
\begin{equation}
\centering
\dot{c}(t) = - \frac{c}{\tau _{c}} + W(\Delta t) \delta(t - s_{pre/post}) C_{1}	
\end{equation}
where $c$ is an eligibility trace. $s_{pre/post}$ means the time of a pre- or postsynaptic spikes. $C_{1}$ is a constant coefficient. $\tau_{c}$ is a time constant of the eligibility trace.
\textcolor{black}{\mbox{$\delta$} is the Dirac delta function.}
\begin{equation}
\centering
\dot{w}(t) = R(t) \times c(t)	
\end{equation}
where $R(t)$ is the reward signal.  More details on the R-STDP mechanism can be found in~\cite{potjans2010enabling, 10.3389/fncir.2015.00085}.

\begin{figure}[!t]
	\centering
	\includegraphics[width=0.78\textwidth]{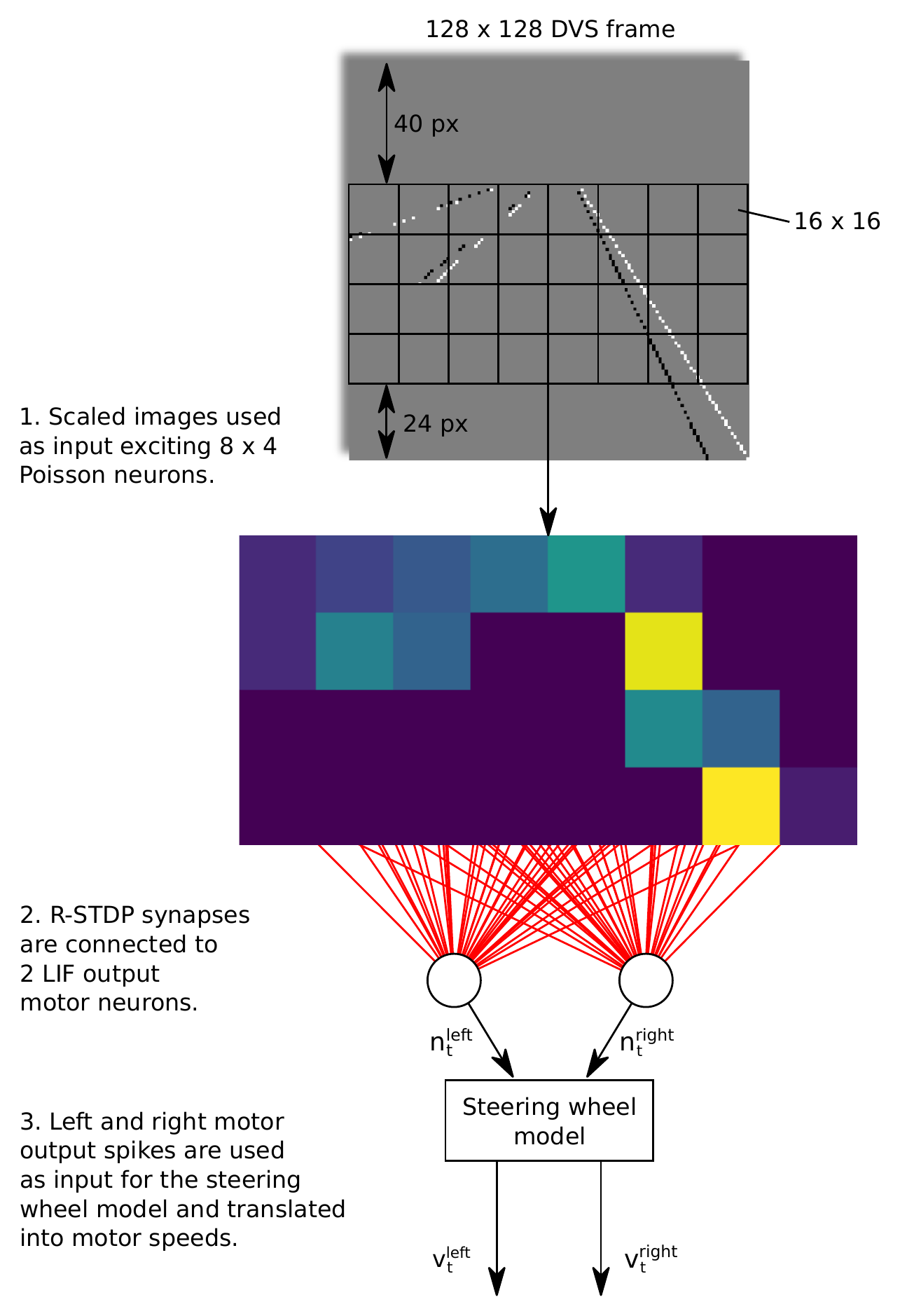}
	\caption[R-STDP network architecture.]{Network architecture of the R-STDP implementation using DVS frames as input.}
	\label{fig:controller_rstdp}
\end{figure}

\subsection{Reward Generation for R-STDP}
\begin{figure}[t]
	\centering
	\includegraphics[width=0.88\textwidth]{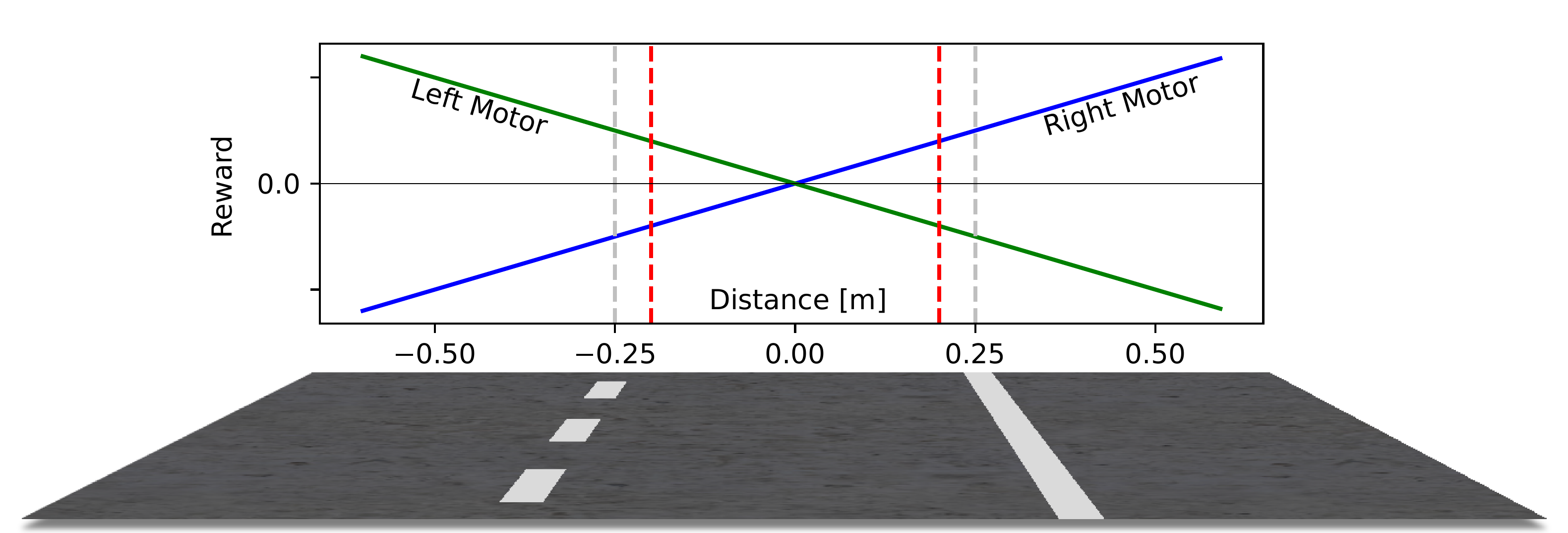}
	\caption[R-STDP reward.]{Reward given by the R-STDP controller: It is defined for each motor individually as a linear function of the lane center distance scaled by a constant $c_r$. The lane markings are $0.25\,m$ away from the lane center. If the robot goes further than $0.2\,m$ from the lane center, episodes are terminated and the robot is positioned at its starting position.}
	\label{fig rstdp reward}
\end{figure}

Instead of dividing the input data and feeding it into two separate networks with static weights as was done in~\cite{kaiser2016towards}, a single SNN based on R-STDP is designed as shown in Fig.~\ref{fig:controller_rstdp}.
The input data is scaled and used for excitation of Poisson neurons, in a single network with $8 \times 4 = 32$ input neurons.
Then, the input layer is connected to two LIF output neurons  in an ``all to all" fashion using R-STDP synapses.
%In order to reduce the complexity involved in the control task, the reward signal is directly set at each simulation time step instead of doing it indirectly by exciting a pool of dopaminergic neurons first.
The reward signal given at each simulation time step is shown in Fig.~\ref{fig rstdp reward}. It is defined for each motor with opposite signs linearly dependent on the robot's distance to the lane center. 
When the robot is on the right side of the lane center and should turn left to get back, connections that lead the right motor neuron to fire are strengthened, connections that lead the left motor neuron to fire are weakened. 
Conversely, if the car is on the opposite side of the lane-center, this process is reversed. 
Over time, the robot should learn to associate certain input stimuli with left or right turns and act accordingly. 
These considerations lead to the following rewards for left and right motor neuron connections, with $d$ being the distance to the lane center and $c_r$ a constant scaling the reward:

\begin{equation}\label{eq rstdp reward}
r_{left/right} = -/+ ( d \cdot c_r )  
\end{equation}

\subsection{Encoding and Decoding}
For communicating with robot sensors and motors in SNNs, the sensory information should be encoded into input spikes and the output spikes should be decoded into motor commands.
A similar processing procedure for the encoding and decoding can be found in~\cite{kaiser2016towards}.  
The same model is implemented in this paper with only one change.
Instead of steering angles, turning speeds are computed and added or subtracted for the left and right motor.
First, the output spike count $n_t^{left(right)}$ is scaled by the maximum possible output $n_{max}$:
\begin{equation}
m_t^{left(right)} = \frac{n_t^{left(right)}}{n_{max}}\in [0;1] \text{,  with } n_{max} = \frac{T_{sim}}{T_{refrac}},
\end{equation}
where $T_{sim}$ denotes the simulation time step length and $T_{refrac}$ describes the refractory period length of the LIF neuron. Based on the difference of the normalized activities $m_t^{left}$ and $m_t^{right}$ and a turning constant $c_{turn}$, the turning speed is defined as
\begin{equation}
S_t = c_{turn} \cdot a_t \text{,   with } a_t = m_t^{left} - m_t^{right} \in [-1;1].
\end{equation}
Furthermore, in order to ensure a minimum running for the robot, the overall speed is controlled according to
\begin{equation}
V_t = - |a_t|\cdot (v_{max} - v_{min}) + v_{max},
\end{equation}
where $v_{min}$ and $v_{max}$ are predefined speed limits. 
Since controlling a car is generally a continuous process, overall speed and turn speed were smoothened based on the activities:
\begin{align}
v_t = c \cdot V_t + (1 - c) \cdot v_{t-1},\\
s_t = c \cdot S_t + (1 - c) \cdot s_{t-1},\\
\text{with } c = \sqrt{ \frac { (m_t^{left} )^2  +  ( m_t^{right} )^2 } {2} }
\end{align}
Finally, the control signals for the left and right motor are computed by
\begin{equation}
v_t^{left} = v_t + s_t \text{ and } v_t^{right} = v_t - s_t.
\end{equation}

\subsection{Training}

In order to train the network successfully, the parameters of the R-STDP controller have to be carefully chosen (see Tab.~\ref{tab:param_rstdp} in the appendix). 
First, the training result is closely related to the reward in~\ref{eq rstdp reward}.
If the value is too low, the learning will take too much time and it might be difficult to see any progress at all. 
By contrast, if it is too high, the learning will become increasingly instable and the robot will not learn anything.
Second, the initial network weights are critical for learning as well. 
In this study, weights are initialized uniformly at a relatively low value of 200. 
The weights have to be larger than zero, because both motor neurons must be excited from the beginning in order to induce weight changes following the R-STDP learning rule. 
%In the best case the initial weight values are as close as possible to their final values after learning. 
%Therefore, the initial weight value has been set to an estimated mean value of the weights after learning. 
Furthermore, the weights are clipped to $[0:3000]$, only allowing excitatory synaptic connections.

\section{Results}
\label{sec:discuss}

With the lane-keeping tasks in mind, DQN, DQN-SNN, and R-STDP controllers for the Pioneer robot were presented earlier in this paper with regard to the basic principles and implementation details. 
In this section, the training results of each controller are discussed and their performances are also compared with each other and with the Braitenberg controller from~\cite{kaiser2016towards}.

\subsection{DQN Training Results}

\begin{figure}[!t]
	\centering
		\includegraphics[width=0.7\linewidth]{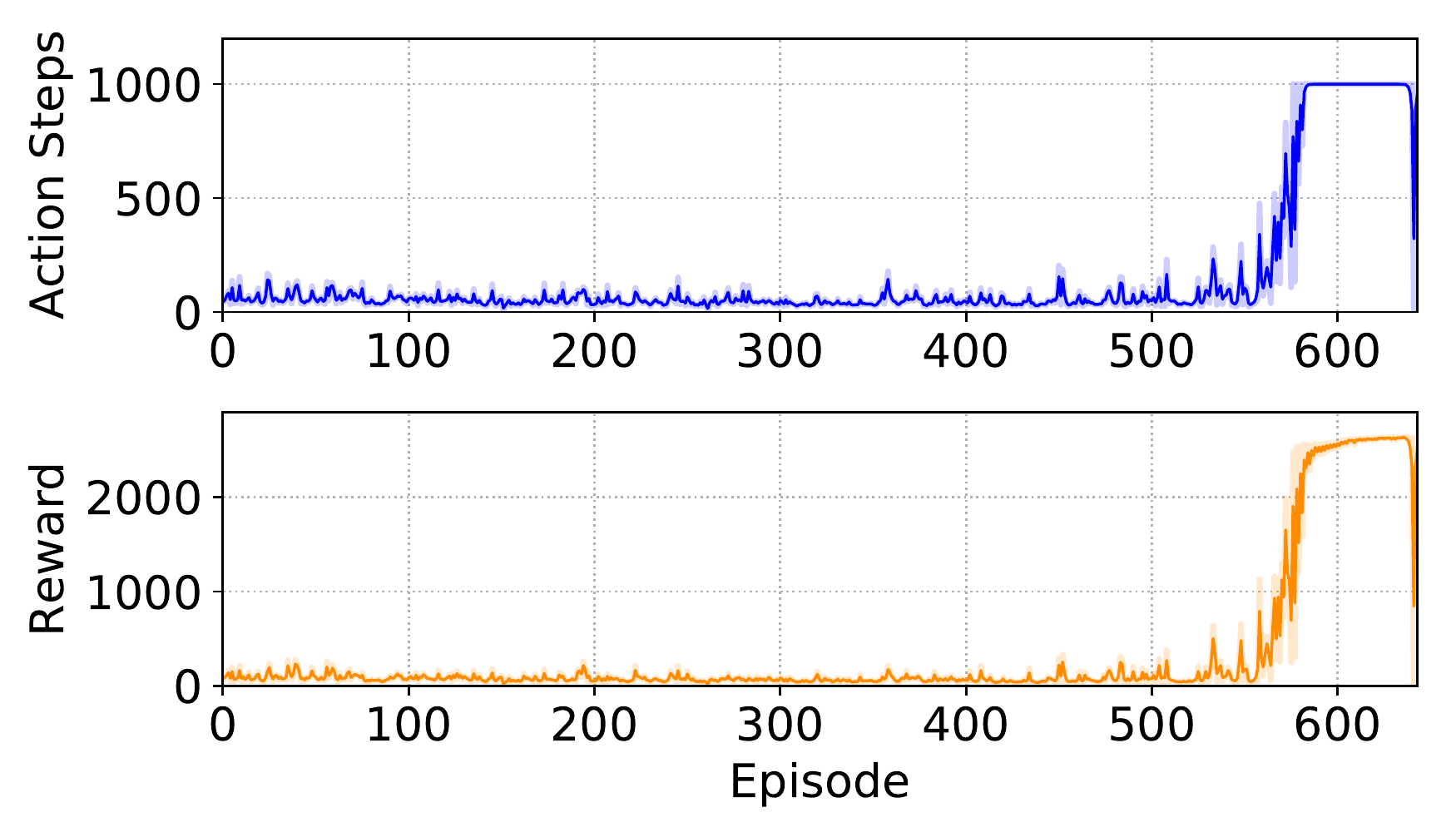}
		\caption{Scenario 1: After 500 episodes, the robot learns to follow the lane without triggering a reset. Episodes are limited to $1{,}000$ action steps. The accumulated rewards collected in $1{,}000$ action steps still improve after the robot has reached the step limit.}
		\label{fig dqn training s1}
\end{figure}

\begin{figure}[!t]
	\centering
		\includegraphics[width=0.7\linewidth]{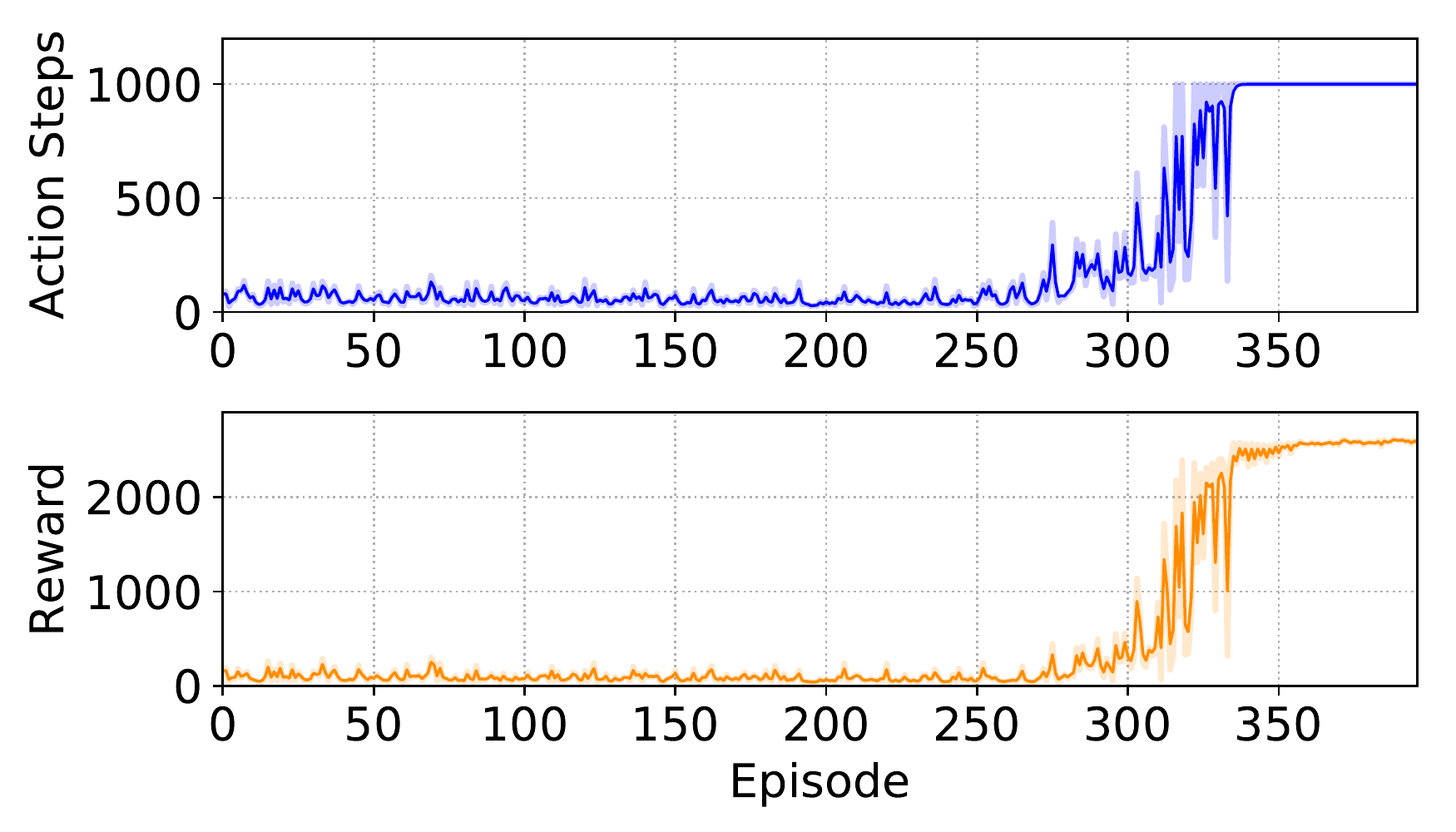}
		\caption{Scenario 2: Action steps and rewards for each training episode of the DQN controller.}
		\label{fig dqn training s2}
\end{figure}

\begin{figure}[!t]
	\centering
		\includegraphics[width=0.7\linewidth]{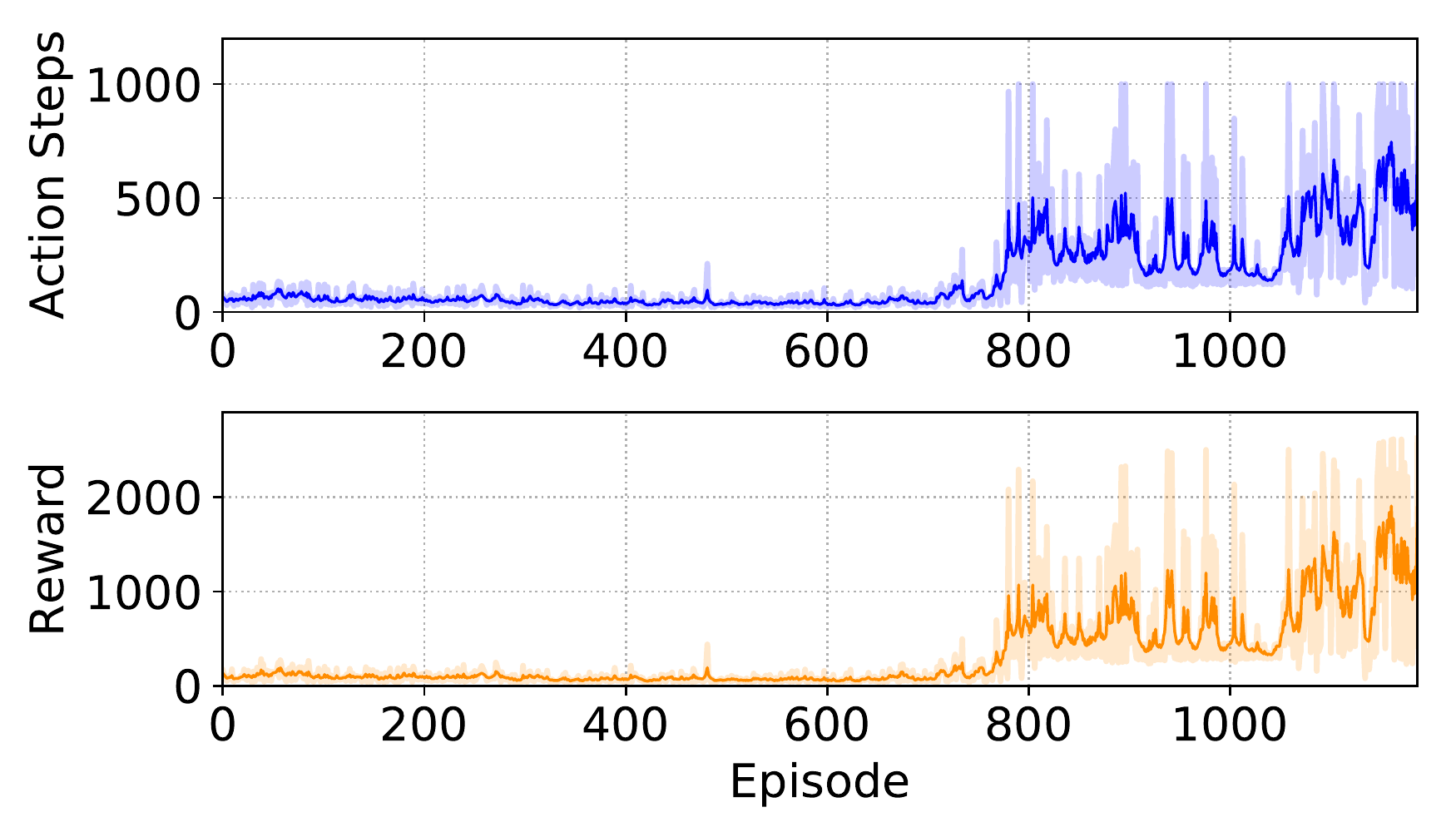}
		\caption{Scenario 3: The algorithm failed to learn a stable policy after $1{,}186$ episodes and $170{,}000$ steps.}
		\label{fig dqn training s3}
\end{figure}

Fig.~\ref{fig dqn training s1} shows the training progress of the DQN algorithm in the first scenario. 
Training parameters are presented in Tab.~\ref{tab param dqn}. 
At the beginning, the robot will randomly choose actions regardless of the rewards. 
Episodes are terminated once these random actions lead the robot beyond the $0.5\,m$ lane-center distance threshold. 
Therefore, action steps and rewards are randomly distributed at a low level at the beginning. 
Even though the $\epsilon$-greedy policy constantly increases the chance of choosing the action with the highest action value, the robot does not show any learning effect until episode 400. 
At around episode 300, action steps and rewards actually decrease, because the robot is following a policy that is not optimal yet. 
After approximately 580 episodes, the robot has learned to follow the lane without trigging a reset. 
To ensure experiences from both inner and outer lanes, even if the robot has successfully learned to follow them, episodes are also terminated after $1{,}000$ action steps ($10{,}000$ time steps), since competing a full lap takes around $5{,}000$ time steps at the pre-defined motor speed. 
After each episode, the robot is placed at the start point of the other side of the road. 
The accumulative rewards have exceeded $1{,}000$ action steps after 580 episodes, and still slowly increase, approaching a reward maximum afterwards.

Similarly, the algorithm learns a control strategy in the second scenario as well. Due to the reduced complexity in the state images, effective learning already begins after 300 episodes (Fig. \ref{fig dqn training s2}). 
Interestingly, we can observe that the DQN learns faster compared with the process in the first scenario.
The reason is that the right side of the input image does not generate any information due to the missing boundary lane, which leads to less complexity in the network.
In the third scenario, by contrast, the DQN algorithm failed to learn a stable policy. 
Fig.~\ref{fig dqn training s3} shows the episode lengths and rewards in $170{,}000$ time steps total. 
The starting positions in the scenario are switched so that the robot experiences both road patterns from the beginning. 
At around episode 800, the robot learns to follow the lane, even completing laps and reaching the time step limit at times. 
Unfortunately, it does not learn a generalized policy that works for both lanes. 
Once the algorithm figures out how to take a turn or a transition section from one pattern to another, it seems to have detrimental effects on its behavior in other situations. 
Even though the average reward over several episodes increases towards the end, the algorithm never reaches the time step limit in consecutive episodes. 
Taken together, the algorithm in the third scenario optimizes its behavior but fails to reach a global reward maximum.

%\begin{figure}[t]
%	\centering
%	\includegraphics[width=0.48\textwidth]{figures/rstdp_training.pdf}
%	\caption[R-STDP training.]{Scenario 1. Learning progress of the R-STDP controller. The termination position when the robot causes a reset and the network weights are shown over the number of simulation steps ($1\, step = 50\, ms$). During the first 10000 simulation steps, the robot causes resets at each trial in the first turn in both directions (Termination in section B). Afterwards, it has successfully learned how to follow the lane only causing a reset when a complete lap is finished.}
%	\label{fig rstdp training s1}
%\end{figure}
%\begin{figure}[t]
%	\centering
%	\includegraphics[width=0.45\textwidth]{figures/rstdp_weights.pdf}
%	\caption[R-STDP weights.]{Scenario 1. Learned connection weights to the left and right motor neuron of the R-STDP controller after 30000 simulation steps.}
%	\label{fig rstdp weights s1}
%\end{figure}

\subsection{DQN-SNN Training Results}

For the policy transfer from the DQN action network to an SNN, an \textcolor{black}{state-action dataset} is created using the state samples stored in the experience buffer. 
During training of the DQN controller in the first scenario, $98{,}990$ state samples are visited and stored. 
At the beginning of the training procedure, the robot experiences many states that are far from the optimal lane center position. 
Once it has learned to follow the lane, however, it will experience only states close the lane center. 
These states are much more important for the policy transfer, because the robot controlled by the SNN will likely never see those ``poor" states far from the lane center. 
Therefore, it is important to train the SNN on a dataset with mostly ``good" states. 
This is achieved by letting the robot run and collect states for a while after successfully learning a good policy to ensure a favorable distribution of states in the dataset. 
Using the previously trained DQN action network, all states are labeled with actions and an ANN is trained reaching a classification accuracy of $93.05\%$. 
Further training and network parameters are shown in Tab.~\ref{tab param dqn-snn}. 
Following that, the network weights are normalized and transferred to an SNN based on Algorithm~\ref{fig snn alg} with the same architecture performing the robot control task. 
In the second scenario, $100{,}236$ states could be classified with an accuracy of 91.71\% following the same procedure. 
Due to the fact that the DQN algorithm could not successfully learn a stable policy in the third scenario, the DQN-SNN controller is only implemented for the first two scenarios.

\subsection{Braitenberg Vehicle Controller}
\begin{figure}[t]
	\centering
	\includegraphics[width=0.7\textwidth]{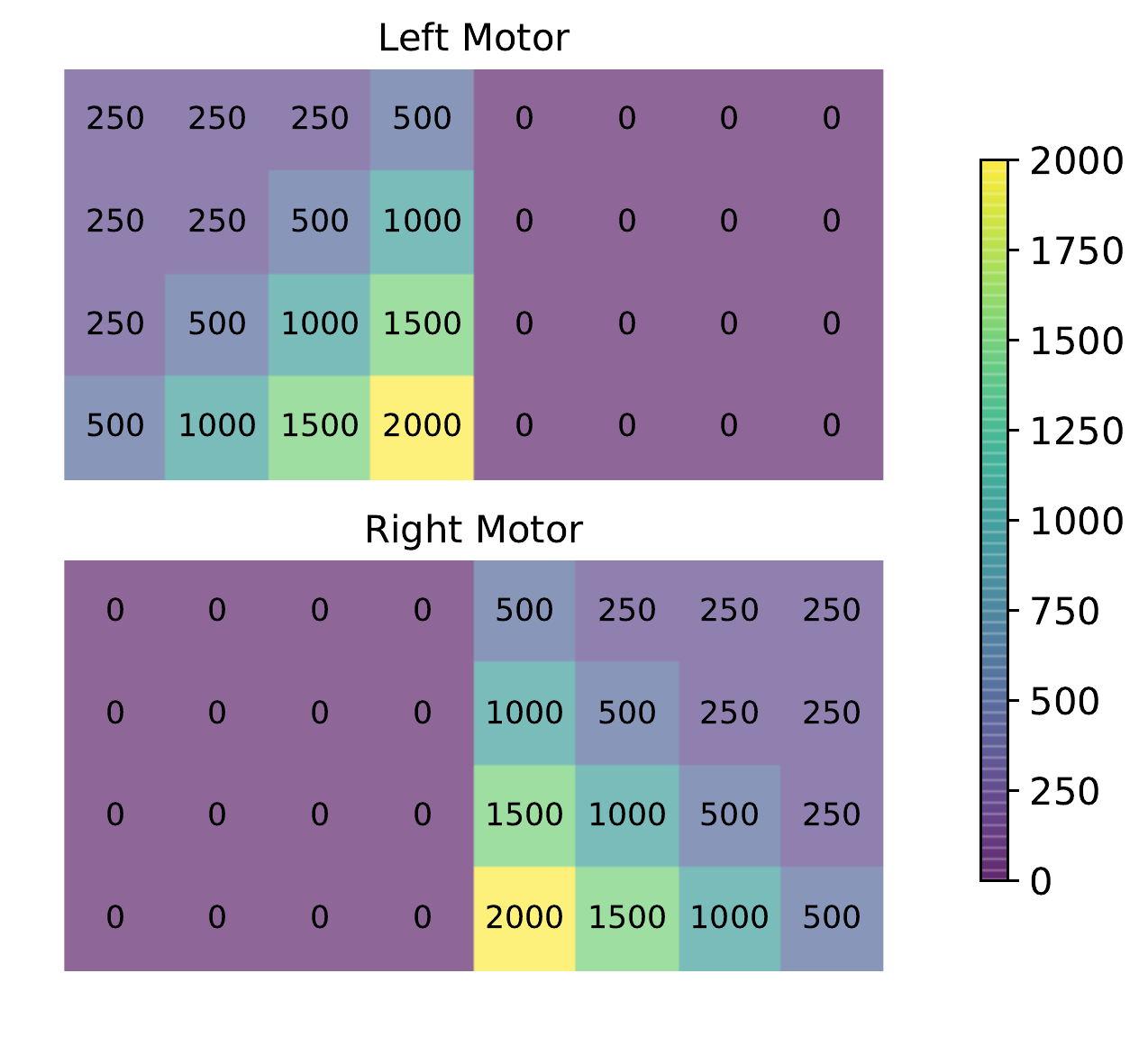}
	\caption[Scenario 1: Braitenberg network weights.]{Scenario 1: Static connection weights to the left and right motor neuron of the Braitenberg vehicle controller.}
	\label{fig braitenberg weights s1}
\end{figure}
To serve as a basis for further investigations, Kaiser proposed a simple Braitenberg vehicle controller for the lane following task~\cite{kaiser2016towards}. 
Depending on simple static connection schemes between sensors and motors, the vehicle exhibits simple animal-like behavior, such as turning towards or away from a sensory stimulus, e.g. in the form of light.

\begin{figure*}[!t]
	\centering
	\includegraphics[angle=0, width=0.95\textwidth]{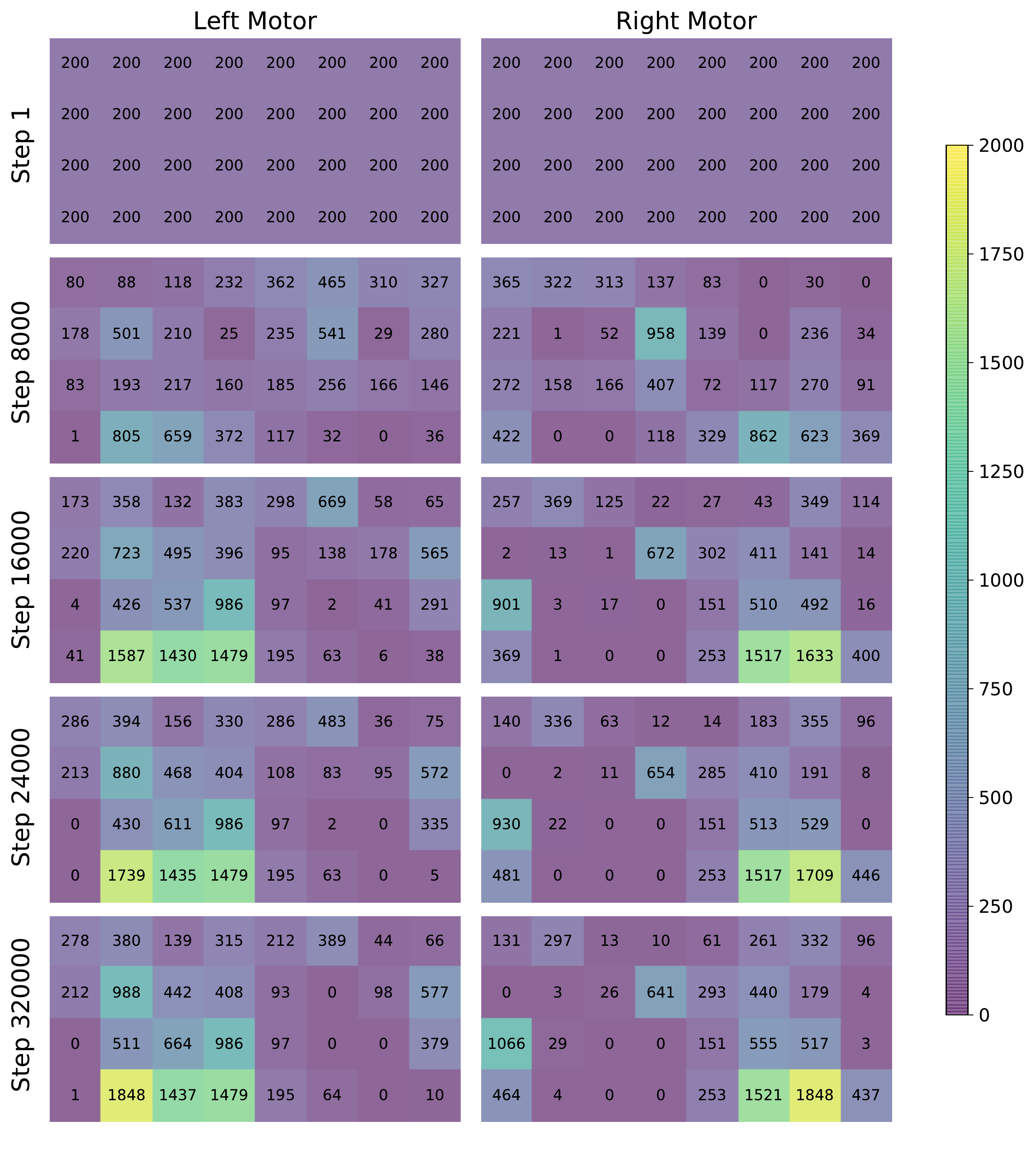}
	\caption{Scenario 1. Learning progress of the R-STDP controller over every $8{,}000$ steps ($1\, step = 50\, ms$). 
		Learned connection weights to the left and right motor neuron of the R-STDP controller are shown in last row after $30,000$ simulation steps.
	}
	\label{fig rstdp training s1}
\end{figure*}
\begin{figure}[t]
	\centering
	\includegraphics[angle=0, width=0.78\textwidth]{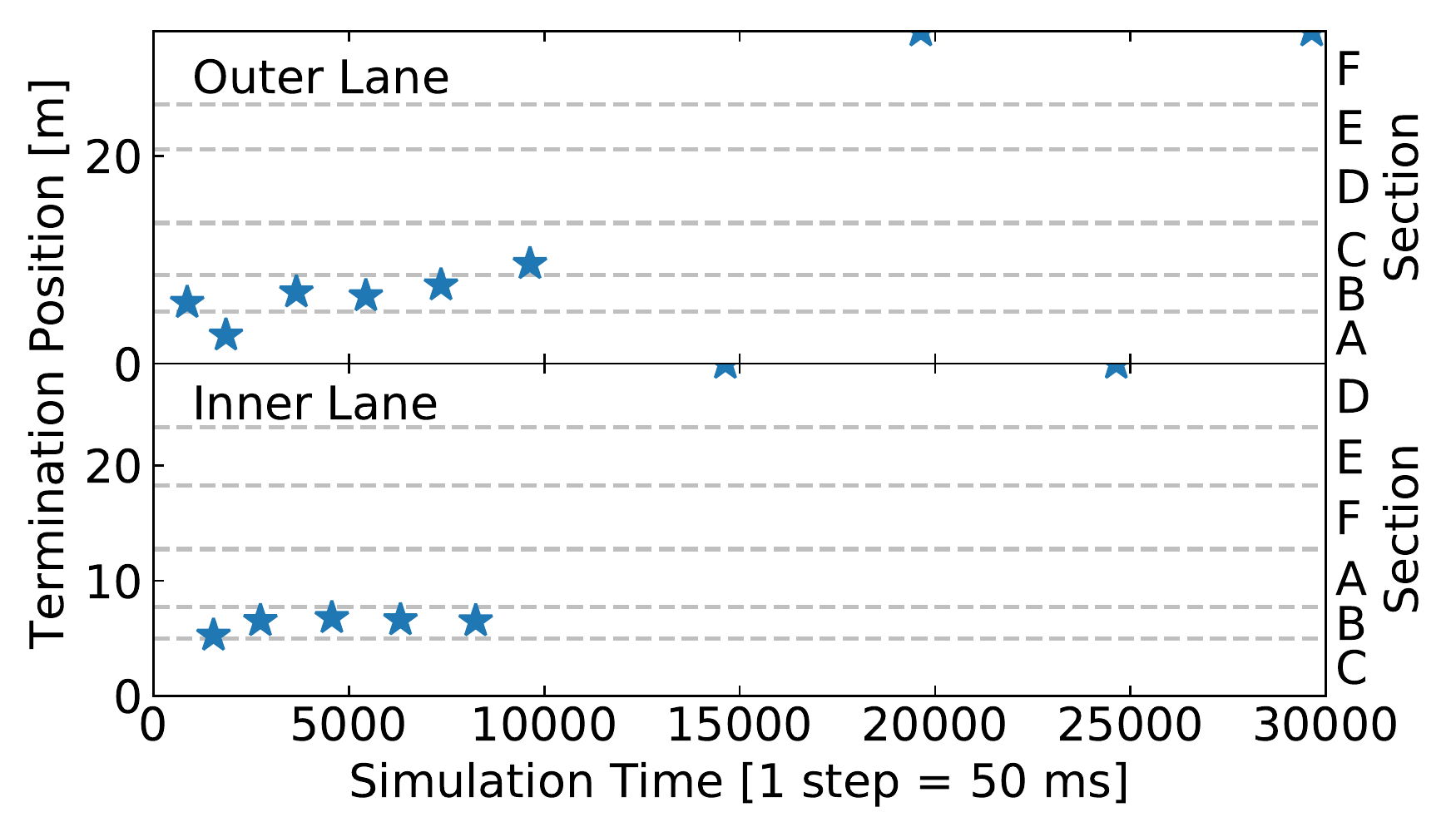}
	\caption{Scenario 1. Termination position of the robot at each trail is marked by a star. 
		During the first $10{,}000$ simulation steps, the robot triggers resets at each trial in the first turn in both directions. Afterwards, it has successfully learned how to follow the lane, only triggering a reset when a complete lap is finished.}
	\label{fig rstdp termination s1}
\end{figure}

In a classic Braitenberg vehicle, the activity of sensory inputs steers the agent towards stimuli or away from stimuli depending on the connection scheme. 
In the first scenario, the robot is supposed to follow the lane without crossing the solid line on the right or the dashed line in the middle of the road. 
Therefore, if the robot deviates from the lane center, the motor neuron activities should increase or decrease so that the robot adjusts its direction accordingly. 
Fig.~\ref{fig braitenberg weights s1} shows the weights of the synaptic connections to the left and right motor neuron. 
If a line in the robot's vision gets closer to the bottom center of the image, the related motor neuron activity will be increased while the opposite motor neuron's activity will be decreased. 
If the robot gets close to the solid line on its right side, for example, left and right motor neurons will decrease and increase their firing rate, respectively, causing the robot to turn to the left. 
The same principle applies for the opposite side as well. 
The network weights are chosen manually by trial and error. 
This controller is only applied in the first scenario for further performance comparison.

%\begin{figure}[t]
%	\centering
%	\includegraphics[width=0.45\textwidth]{figures/rstdp_training_2.pdf}
%	\caption[R-STDP training.]{The termination position when the robot causes a reset and the network weights are shown over the number of simulation steps (1 step = 50ms)}
%	\label{fig rstdp training s2}
%\end{figure}
%\begin{figure}[h]
%	\centering
%	\includegraphics[width=0.45\textwidth]{figures/rstdp_weights_2.pdf}
%	\caption[R-STDP weights.]{Scenario 2. Learned connection weights to the left and right motor neuron of the R-STDP controller after 30000 simulation steps.}
%	\label{fig rstdp weights s2}
%\end{figure} 

\subsection{R-STDP Training Results}

%\vspace{-10 mm}
\begin{figure*}[!t]
	\centering
	\includegraphics[width=0.95\textwidth]{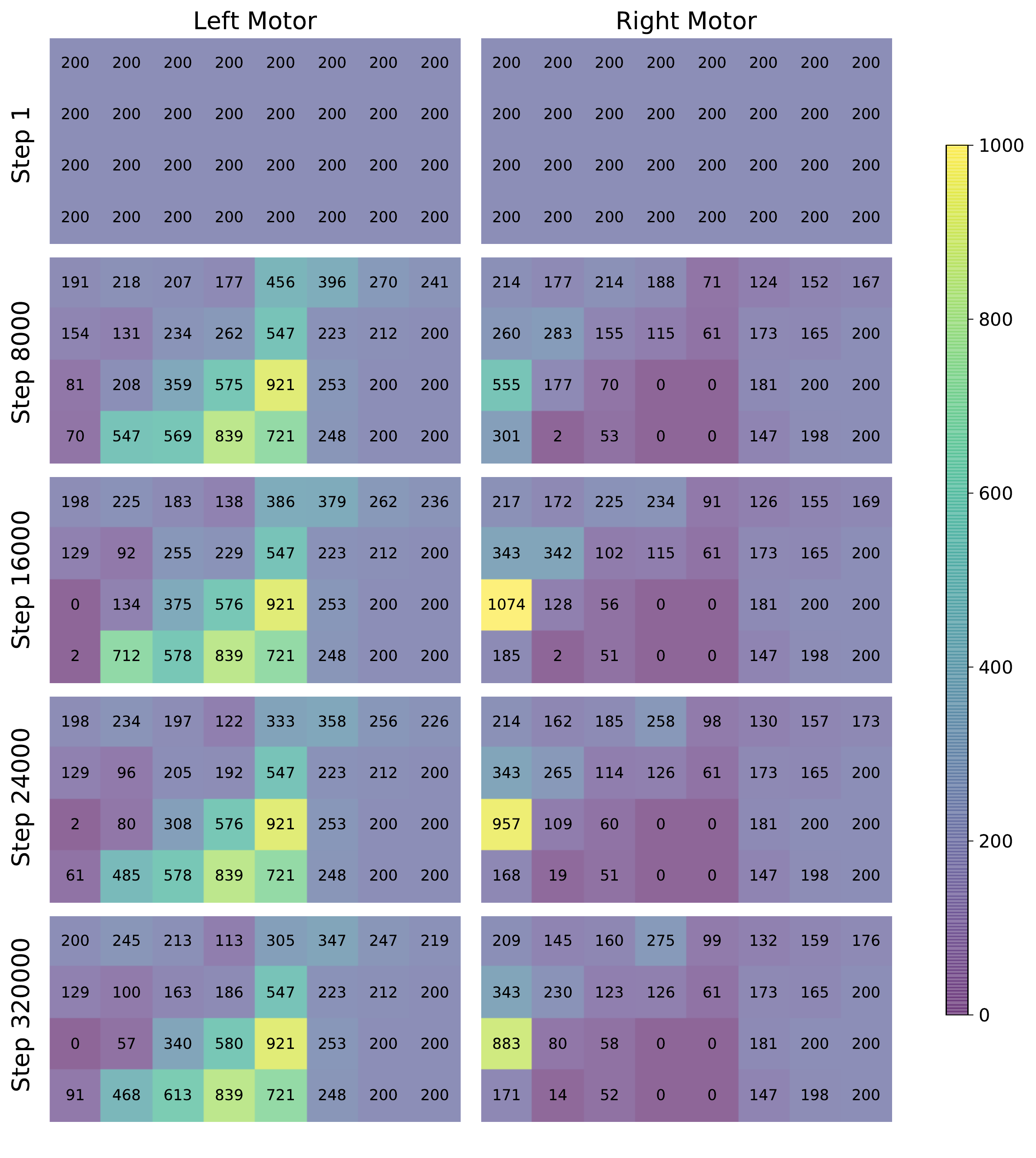}
	\caption[Scenario 2: Braitenberg network weights.]{Scenario 2. Learning progress of the R-STDP controller over every $8{,}000$ steps ($1\, step = 50\, ms$). 
		Learned connection weights to the left and right motor neuron of the R-STDP controller are shown in last row after $30{,}000$ simulation steps.
		}
	\label{fig rstdp training s2}
\end{figure*}
\begin{figure}[t]
	\centering
	\includegraphics[angle=0, width=0.78\textwidth]{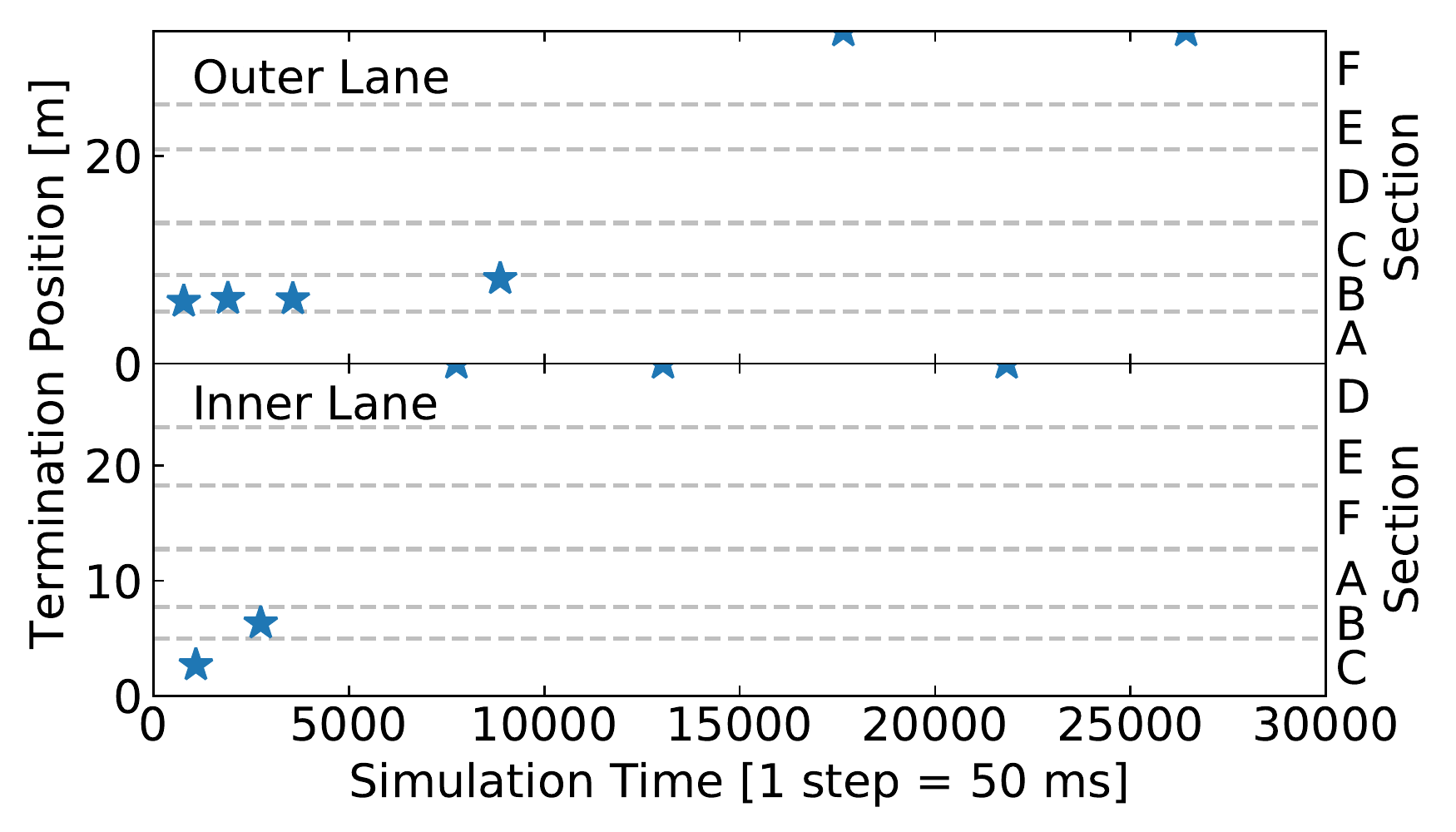}
	\caption{Scenario 2. Termination position of the robot at each trail is marked by a star. 
	The robot is mostly reset in sections B until laps are completed on both lanes after approximately $14{,}000$ steps.}
	\label{fig rstdp termination s2}
\end{figure}
\begin{figure*}[t!]
	\centering
	\includegraphics[angle=0, width=0.95\textwidth]{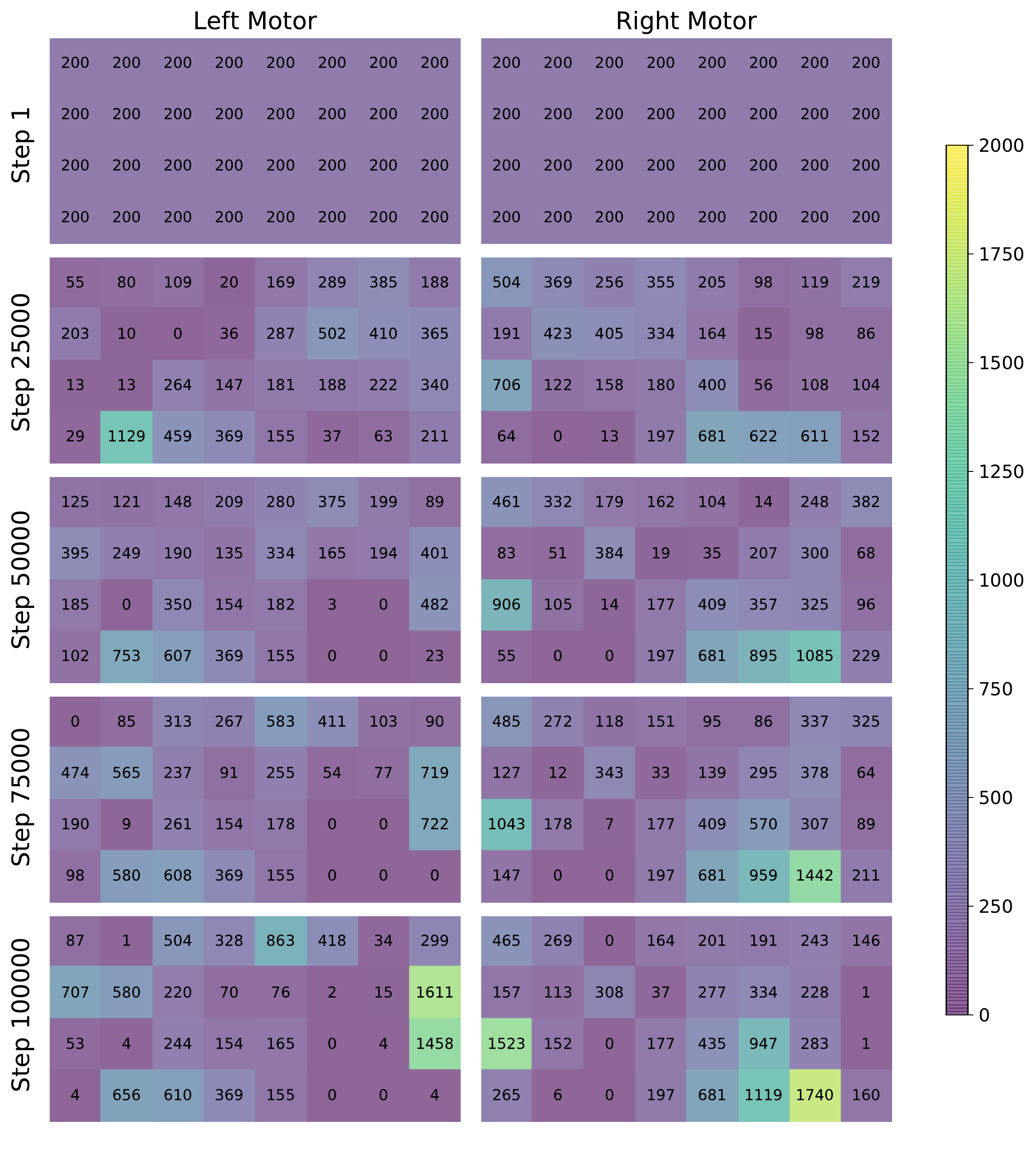}
	\caption[Scenario 3: Braitenberg network weights.]{Scenario 3. Learning progress of the R-STDP controller over every $25{,}000$ steps ($1\, step = 50\, ms$). 
		Learned connection weights to the left and right motor neuron of the R-STDP controller are shown in last row after $100{,}000$ simulation steps.
		}
	\label{fig rstdp training s3}
\end{figure*}
\begin{figure}[t!]
	\centering
	\includegraphics[angle=0, width=0.78\textwidth]{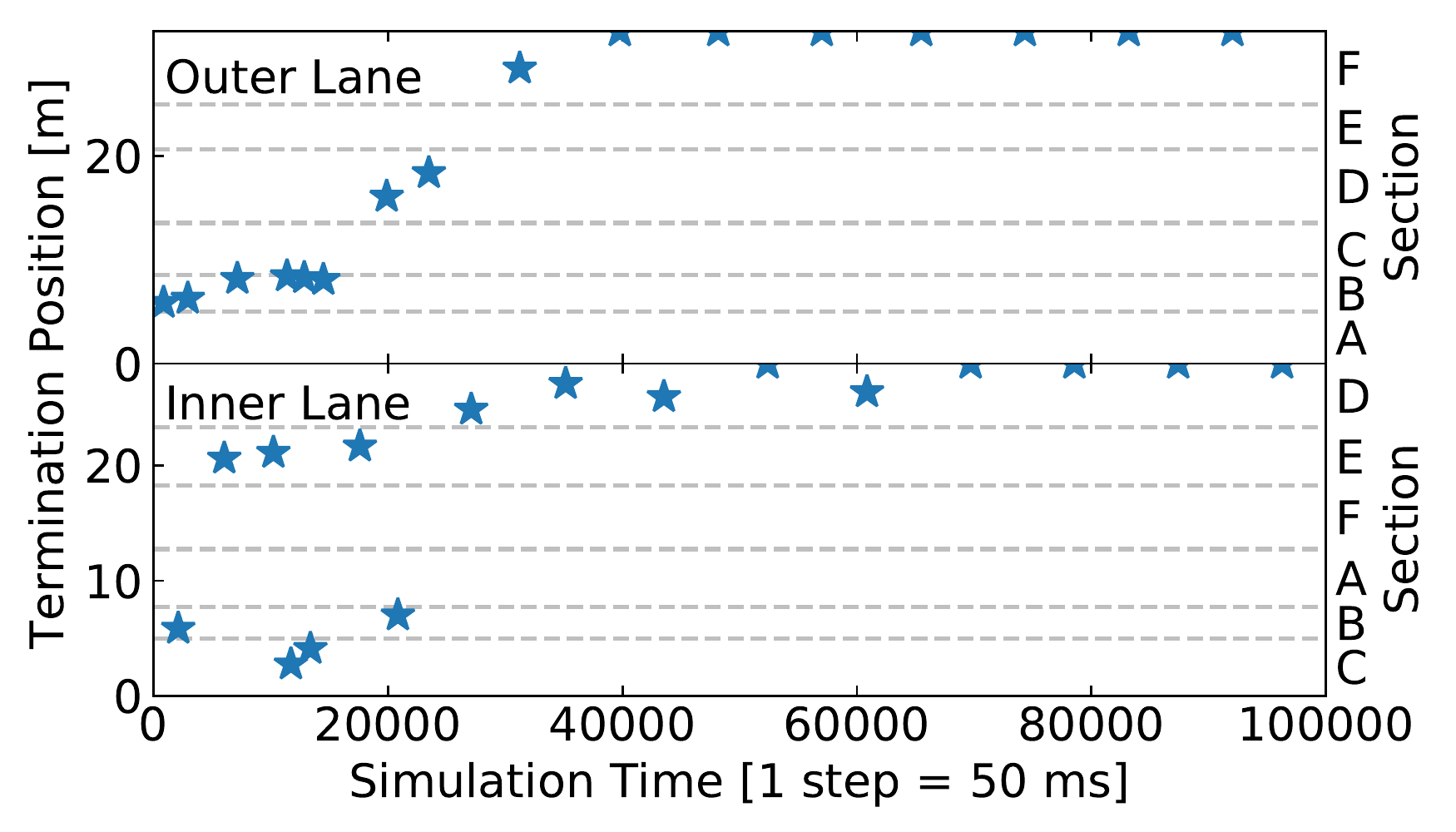}
	\caption{Scenario 3. Termination position of the robot at each trail is marked by a star. After an initial learning phase, the robot is mostly reset in sections B and D until laps are completed on both lanes after approximately $75{,}000$ steps.}
	\label{fig rstdp termination s3}
\end{figure}

\textcolor{black}{\mbox{Fig.~\ref{fig rstdp training s1}} shows the training progress of the R-STDP controller in the first scenario.}
Specifically, the changes in the synaptic weights are shown every $8{,}000$ steps over the course of the simulation.
Fig.~\ref{fig rstdp termination s1} shows the termination position of the robot at each trail when it exceeds the lane center distance of $0.2~m$, triggering a reset.
A simulation step is equivalent to $50~ms$ both for the simulation of the SNN as well as the robot simulator itself.
At the beginning of the training procedure, the robot will go straight forward, because all connection weights for both motor neurons have been set to the same value.
Therefore, during the first $10{,}000$ simulation steps, trials are mostly terminated at the first turn in both directions, when the robot misses the turn and the lane center distance exceeds $0.2~m$.
Each time the robot misses a turn, it will periodically induce high reward values at the beginning, changing the synaptic weights.
Shortly before step $10{,}000$, the robot has learned to take the turn, but it still deviates from the optimal lane center position.
Consequently, the high reward over a longer period of time causes a significant change in the connection values.
The learned weights after $30{,}000$ simulation steps are shown in the last row of Fig.~\ref{fig rstdp training s1}.
Interestingly, the connection weights resemble the theoretically derived weights of the Braitenberg controller (see Fig.~\ref{fig braitenberg weights s1}), with very low values in one half of the image and increasing values from the top corner to the bottom center in the other half of the image.
Furthermore, it can be seen that left and right motor neurons mostly seem to be triggered  through the middle and right road line enclosing the lane.
%\subsection{Different Task Scenarios}
%To examine the practicality of the proposed algorithm, another two lane scenarios are implemented (See Fig.~\ref{fig:scenario_sub2} and Fig.~\ref{Lanes with two different patterns}). 

The training results of the controller in the second scenario are shown in Fig.~\ref{fig rstdp training s2} and Fig.~\ref{fig rstdp termination s2}.
The results are similar to the first scenario, completing the first full lap in less than $5{,}000$ simulation steps.
The weights of the controller network after $30{,}000$ simulation steps are shown in the last row of Fig.~\ref{fig rstdp training s2}. 
While the networks weights on the left side from both motor neurons resemble the connection weights learned in the first scenario, it can easily be seen that the weights on the right side have been left unchanged, due to the missing lines in this scenario and the consequential lack of activity during training.

Fig.~\ref{fig rstdp training s3} shows the learning progress during the training in the third scenario. 
First, learning a successful control strategy takes considerably more time than in the first two scenarios.
The obvious explanation for this is that the third scenario incorporates two different road patterns, making the environment more complicated.
Therefore, the controller has to distinguish between a higher number of different situations as well as slowing down the learning procedure.
Moreover, due to the simple fact that the robot does not encounter certain situations until it has learned how to get there, it will only start learning a generalized control strategy that works for both lanes towards the end.
%In the left motor plot it can be seen that some weights might even be increased in the beginning and decreased again afterwards.
The termination positions are shown in Fig.~\ref{fig rstdp termination s3}. As we can see,
after an initial learning phase until approximately step $20{,}000$, the controller is mostly reset in section B (outer lane) and D (inner lane).
When the weights have adapted sufficiently after approximately $75{,}000$ steps, the robot finished the laps on both lanes.
The last row of Fig.~\ref{fig rstdp training s3} shows the learned weights after $100{,}000$ steps.
In comparison to the first scenario, the weight patterns seem very similar, which makes sense considering the fact that the road pattern in the first scenario is the combination of both road patterns in the third scenario. 

%\begin{figure}[t]
%	\centering
%	\includegraphics[width=0.45\textwidth]{figures/scenario3_training.pdf}
%	\caption[R-STDP training.]{Scenario 3. After an initial learning phase, the robot is mostly reset in sections B and D until laps are completed on both lanes after approximately 75000 steps.}
%	\label{fig rstdp training s3}
%\end{figure}
%
%\begin{figure}[t]
%	\centering
%	\includegraphics[width=0.45\textwidth]{figures/rstdp_weights_3.pdf}
%	\caption[R-STDP weights.]{Scenario 3. Learned connection weights to the left and right motor neuron of the R-STDP controller after 100000 simulation steps.}
%	\label{fig rstdp weights s3}
%\end{figure}

\section{Performance \& Comparison}
\label{sec:performance}

\begin{figure*}[t]
	\centering
	\includegraphics[width=1.05\textwidth]{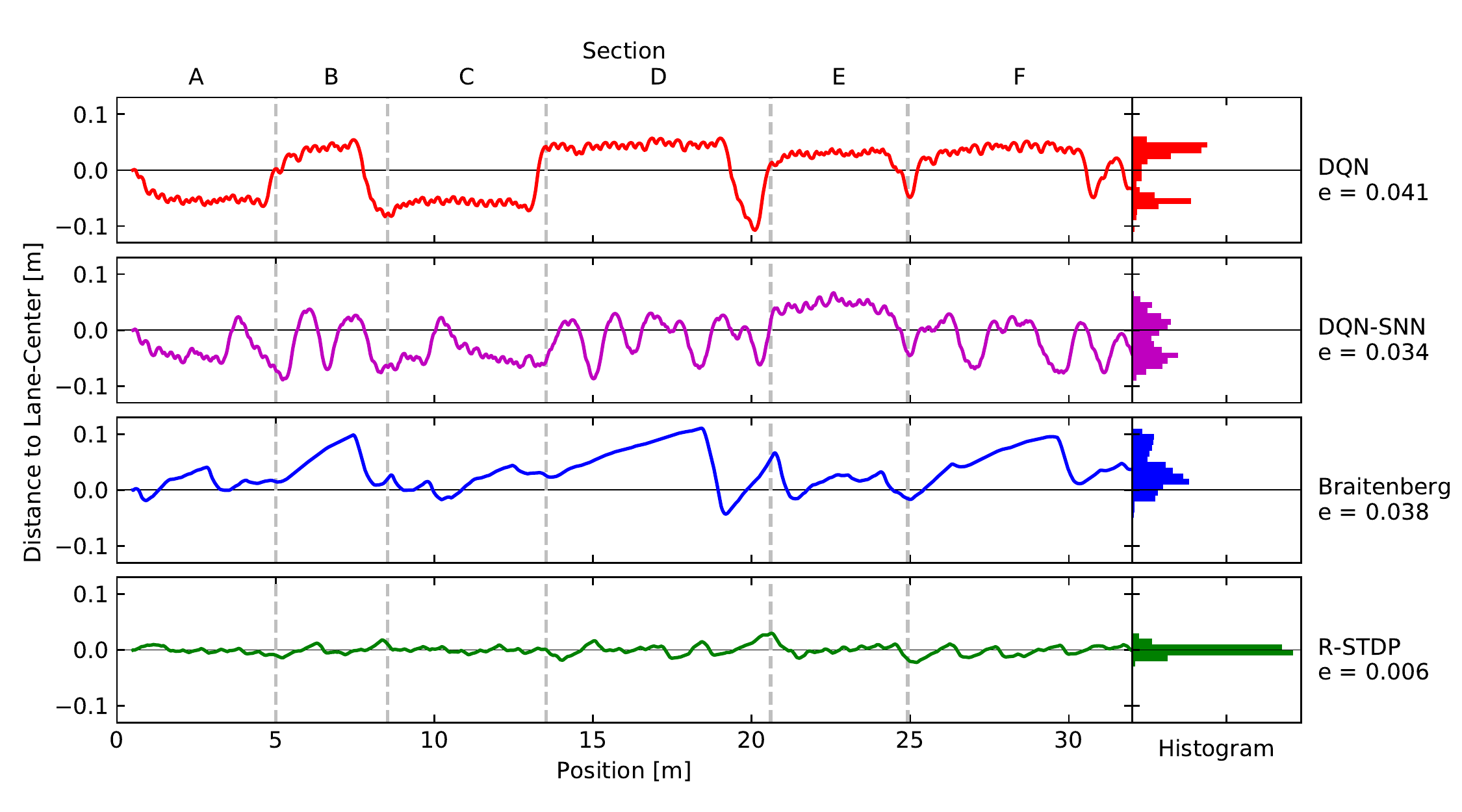}
	\caption[Controller performance in scenario 1.]{Scenario 1: Comparison of different controllers on the outer lane. The deviation from the lane center is shown over the robot position projected to the lane center. Positive lane center distances correspond to deviations to the right side, negative distances to the left side. Course sections are marked by vertical dashed lines (A=straight, B=left, C=straight, D=left, E=right, F=left). On the right side, error distributions for all controllers as well as mean errors $e$ (mean distance to the lane center) are shown.}
	\label{fig performance s1}
\end{figure*}

\begin{figure*}[t]
	\centering
	\includegraphics[width=1.05\textwidth]{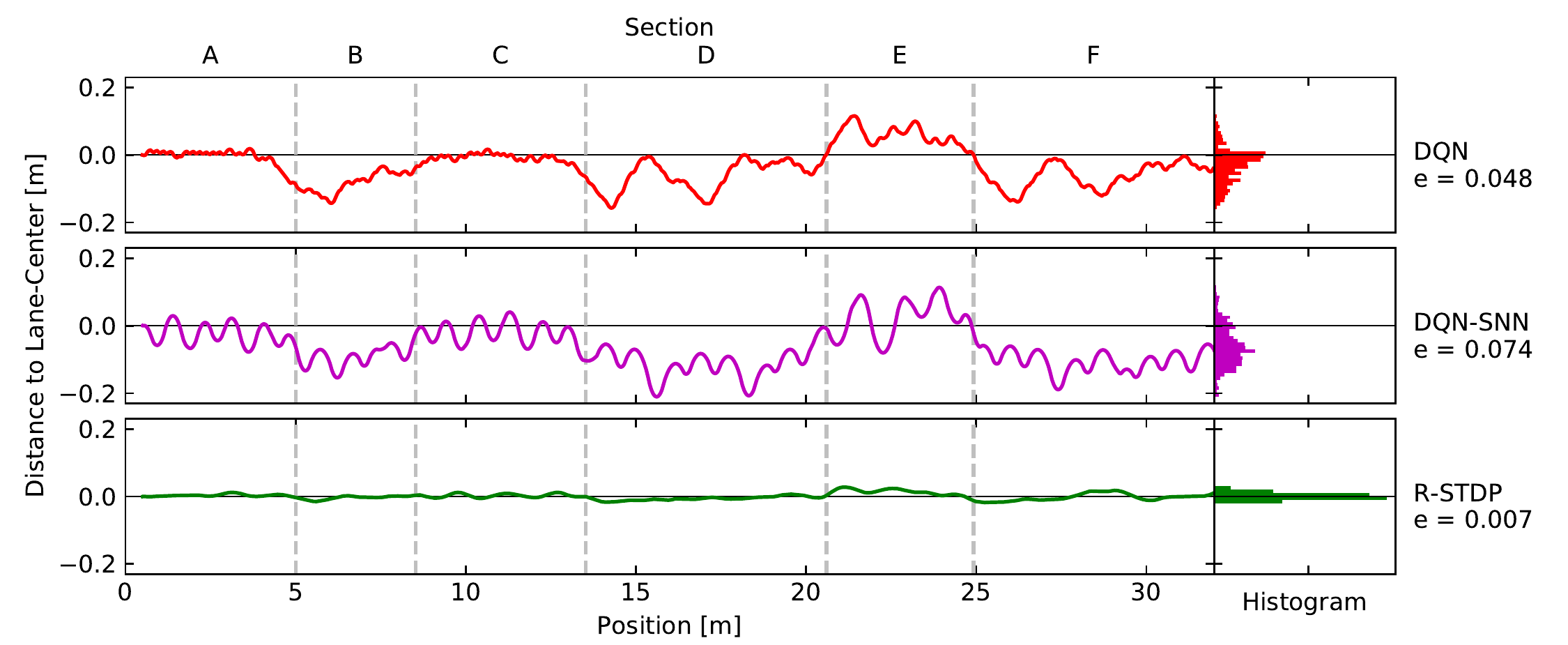}
	\caption[Controller performance in scenario 2.]{Scenario 2: Comparison of different controllers on the outer lane. The deviation from the lane center is shown over the robot position projected to the lane center. Positive lane center distances correspond to deviations to the right side, negative distances to the left side. Course sections are marked by vertical dashed lines (A=straight, B=left, C=straight, D=left, E=right, F=left). On the right side, error distributions for all controllers as well as mean errors $e$ (mean distance to the lane center) are shown.}
	\label{fig performance s2}
\end{figure*}

\begin{figure*}[t]
	\centering
	\includegraphics[width=1.05\textwidth]{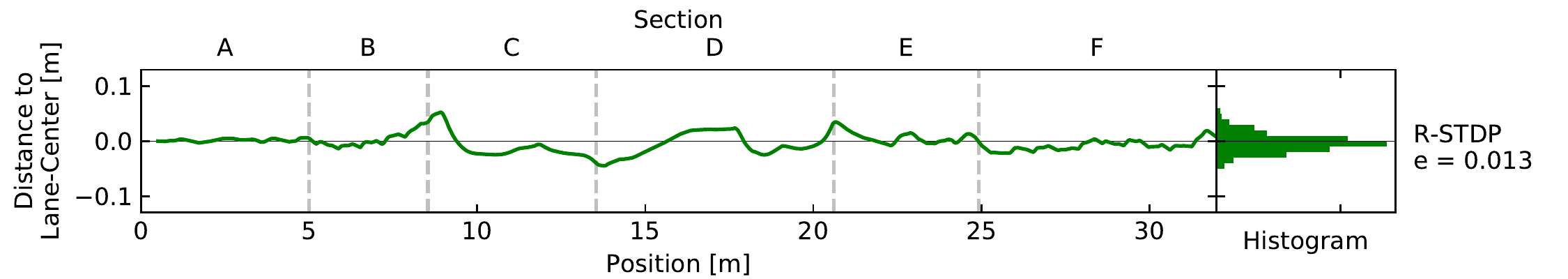}
	\caption[Controller performance in scenario 3.]{Scenario 3: R-STDP controller on the outer lane. The deviation from the lane center is shown over the robot position projected to the lane-center. Positive lane center distances correspond to deviations to the right side, negative distances to the left side. Course sections are marked by vertical dashed lines (A=straight, B=left, C=straight, D=left, E=right, F=left). On the right side, the error distribution as well as the mean error $e$ (mean distance to the lane center) are shown.}
	\label{fig performance s3}
\end{figure*}

At the beginning, all the controllers are successfully trained and tested in the first scenario. 
While the Braitenberg controller is only implemented for the first scenario for comparison purposes, the remaining controllers learned a control strategy for the second scenario as well. 
Only the R-STDP controller, however, learned a stable control strategy for the third scenario. 
In order to obtain comparable performance metrics for each controller, they are evaluated after completing one lap on the outer lane in each scenario. 
Figs. \ref{fig performance s1} - \ref{fig performance s3} show the deviation of the robot from the lane center over the projected course position during one lap for each successful controller in all three scenarios, respectively. 
Moreover, the course is divided into the six sections as shown in Fig. \ref{fig:scenario_sub1}. 
The robot path representation as a projection to the lane center line allows for a numerical analysis of the performance of all controllers . 
Specifically, the error distribution (distance to the lane center) can be shown in the form of a histogram as well as the mean error for each controller.

Initially, the DQN controller is trained and tested in the first scenario (See Fig.~\ref{fig performance s1}). 
The behavior of the robot very clearly depends on the section that it is in. In the straight sections A and C of the first scenario, the robot exhibits a tendency to the left ($-0.1$, defined in Fig~\ref{fig dqn reward}). 
In the left turn sections B, D, and F as well as the right turn section E, the robot tends to the right side of the lane ($+0.1$). 
While the controller does not optimally minimize the lane center distance over the whole course, it seems to be very stable with a constant deviation during each section. 
This behavior can also be seen in the error histogram with two peaks at both sides of the lane center. 
In contrast to the other controllers, the DQN algorithm leads to the highest numerical error with a mean deviation of $e = 0.041\,m$. 
%\textcolor{red}{Interestingly, even though the MDP was defined with 3 discrete actions \emph{left}, \emph{straight} and \emph{right}, the algorithm only chooses between turning left and right after training.} 
In the second scenario, the DQN algorithm shows a higher mean error (See Fig.~\ref{fig performance s2}), which can be explained by the reduced information in the state images. 
Especially in turns on the outer lane, when the robot sees only a very small part of the dashed line, there are only a few pixels containing any information. 
During the straight sections A and C, on the other hand, the robot follows the lane very close to its center, having enough information for a near-optimal control strategy. 
As discussed in the previous section, the DQN algorithm does not learn a stable policy in the third scenario that combines two different road patterns. 
During training, however, it manages to complete full laps and reaches the time step limit several times, proving that it can in fact learn a good policy with different road patterns. 
The problem here seems to be a general policy that works for both lanes, handling the full complexity of the task. 
Considering that other reinforcement learning tasks have successfully been solved using DQN (e.g. playing Atari games in \cite{mnih2015human}), it seems likely that this is mainly due to the simple network architecture that is implemented in this study which fails to evaluate states accurately enough in order to learn a stable policy.

Following the DQN controller, an SNN is trained in order to approximate the policy learned by the DQN algorithm. 
Therefore, when looking at the performed lap, the DQN-SNN controller exhibits some similarities to the DQN controller, e.g. its left tendency in straight sections (A and C) or its right tendency in right turns (E). 
Overall, the controller seems more unstable, exhibiting a lot more oscillatory behavior, especially in left turns (sections B, D and F). 
When looking at the histogram, the error distribution of the DQN-SNN controller looks like a smoothened version of the DQN controller. 
Moreover, the mean error of the transferred SNN controller is surprisingly lower than the one of the original DQN controller. One explanation for this interesting behavior could be the decision frequency that is much higher than before. 
For every decision, the DQN controller collects consecutive frames in order to have enough data and combines them into one single state image. 
In this study, states images are composed of 10 DVS frames and decisions are made every $10 \times 50\,ms = 500 \, ms$. 
The SNN, on the other hand, does not have to accumulate DVS frames beforehand. 
The network architecture will combine the data in the membrane potentials over time. 
Therefore, the network output can be read every $50\,ms$ without having to wait for 10 simulation steps, although it takes some time until enough data has been propagated through the network to produce meaningful output spikes. 
In fact, in many time steps during the simulation the SNN will not produce any output spikes at all, which is why action traces (defined in \ref{equ:trace}) are used to ensure a control signal even if there are no output spikes. 
Considering the loss that was introduced when training the SNN on the state-action dataset, the performance of the controller still seems pretty good. 
In the second scenario (see Fig.~\ref{fig performance s2}), the SNN controller exhibits strong oscillatory behavior. 
Again, this can probably be explained by the reduced amount of information in the image data due to the missing lines. 
If fewer events are created and fed into the network, it takes longer until the information gets propagated through the network and generates an output spike. 
Therefore, the frequency in which the network can make decisions is much lower, resulting in this unstable behavior.

Next, the Braitenberg controller is evaluated while performing the same lap in the first scenario (see Fig.~\ref{fig performance s1}). 
While the controller successfully finishes the course, it can be seen quite clearly that it strongly tends to the right side of the lane, which can be explained by the robot's \textcolor{black}{field of view}. 
In the right half of the DVS images, the robot usually only sees the right solid line. 
In the left half, however, the robot sees the left solid line as well as the dashed middle line of the road, leading to a higher number of detected events and eventually greater activity of the left motor neuron.
This will shift the robot to the right until it has reached a balance in the activity of the motor neurons. 
Even in the right turn (section E), the robot is mostly to the right of the lane center. 
In left turns, the distance to the lane center grows until a point is reached where previously unstimulated neurons with high weights are now excited. 
These will push the right motor neuron activity, leading to a movement correction back to the center. 
This can be seen in all three left turns (sections B, D, and F). 
The value of the controller's mean error during the performed lap is comparable to the first two controllers.

The purpose of the Braitenberg controller is to show the basic underlying control principle here. Instead of improving the controller performance by iteratively adjusting the network weights, the network is re-implemented using R-STDP synapses so that the weights could be automatically learned by the robot. 
In the previous section, we have already shown that those learned weights resemble the theoretically derived weights of the Braitenberg vehicle controller. 
Of all four controllers, the R-STDP controller shows the best performance in this task with comparatively very small deviations from the lane center. 
This gets even clearer when looking at the performance histogram and the mean error that is almost an order of magnitude lower in comparison to the other controllers. 
First, one explanation for this behavior can be found in the very nature of SNNs that allow for high frequency decision-making without the need to split time into discrete steps. 
Second, the R-STDP training algorithm and the related reward are to a great extent tailored to this specific problem. 
The great success of deep reinforcement learning methods such as DQN lies in their capability to learn value functions in high-dimensional state spaces. 
This property allows for a general algorithm that is capable of solving sequential decision-making tasks formulated as MDP, even if rewards are sparse and delayed in time. 
The R-STDP controller, on the other hand, does not have this property. 
Basically, the R-STDP reward can be interpreted as a pre-defined value function with a global maximum that the algorithm will seek out. 
Furthermore, the reward signal incorporates prior knowledge, e.g. that increasing or decreasing motor neuron activities will lead the robot back to the center. 
Therefore, the R-STDP training algorithm solves a mathematically much less complex problem, leaving out the state evaluation step estimating future rewards that is crucial for every classic reinforcement learning algorithm solving MDPs with sparse, delayed rewards.

\begin{figure}[t]
	\centering
	\includegraphics[width=0.7\textwidth]{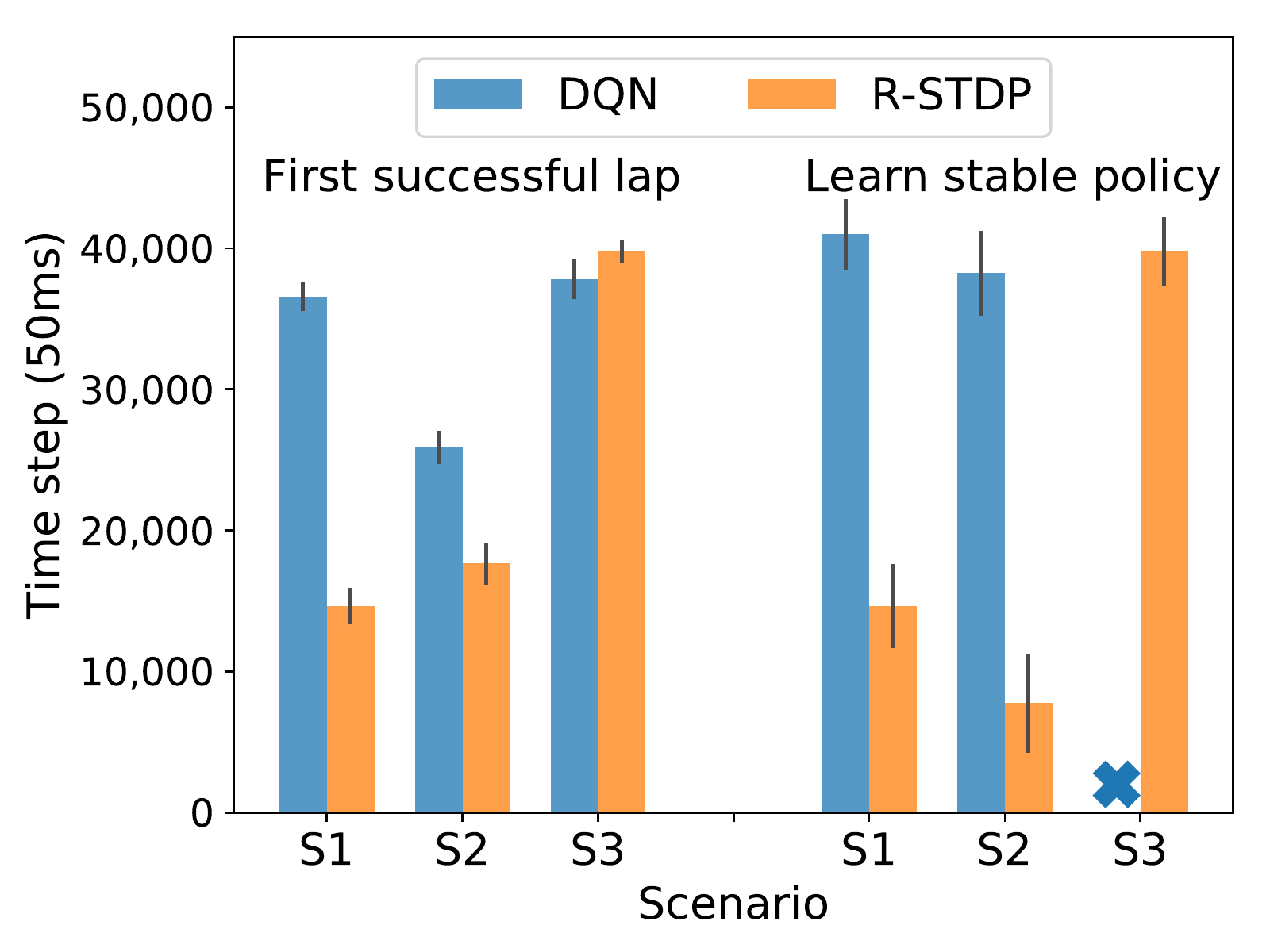}
	\caption{Training time steps comparison for DQN and R-STDP controllers. The left group of bars shows the training time for achieving the first successful lap for both controllers. The right group shows the total training time steps for a stable policy. 
	The standard deviation is marked with a black solid line for three training trails.
	For the DQN controller in the third scenario, it is marked with a $X$, since it fails to learn a stable policy.}
	\label{fig performance time}
\end{figure}

Moreover, the training time steps of the DQN and R-STDP controllers for the three scenarios are shown in Fig.~\ref{fig performance time}. 
For the first two scenarios, the R-STDP controller takes notably less time to learn a stable policy as compared to the DQN controller, even the DQN only takes random actions for the first $1{,}000$ time steps.
For the third scenario, the R-STDP takes about $40{,}000$ time steps to complete the successful learning process, while the DQN fails to learn a stable policy to accomplish consecutive full inner and outer laps.
On the other hand, even for completing the first successful lap, the R-STDP still takes less time.
For the DQN controller, there are still many episodes to be conducted to achieve a stable policy after the first successful trial.
However, for the R-STDP controller, it will almost learn a stable policy once it completes the first successful lap.
The possible explanation for the DQN controller is obvious, since the DQN achieves the first full lap by chance during the process of maximizing its rewards, rather than seeking its global maximum as R-STDP. 
In overall time steps, the R-STDP also beats the DQN, due to its inherent high frequency making for processing event-based data. 

\section{Conclusion}
\label{sec:conclusion}

Spiking neural network, inspired by the mechanism of the brain, offers a promising solution to control robots with biological plausibility and exceptional performances.
However, it lacks sophisticated training algorithms and practical robotic implementations, due to its complexities in constructing and optimizing an SNN.
To bridge this gap, we trained an SNN controller with indirect and direct methods based on DQN policy transfer and R-STDP learning rule, respectively, and further implemented them in lane-keeping tasks for a Pioneer robot. 
For the indirect training, we first trained a DQN controller to accomplish lane-keeping tasks and then transferred its policy to an SNN controller by training it on an state-action dataset using supervised learning.
\textcolor{black}{Our indirect methods offers a quick and efficient way to build up an applicable spike-based controller that is able to be executed on neuromorphic hardwares.}
For the direct training,  
\textcolor{black}{our method directly learns an SNN by utilizing the biological R-STDP learning rule and the event-based vision sensor, aiming to bring reinforcement learning capabilities to SNNs directly.}
Finally, we demonstrated the superiority of the controller trained by R-STDP by comparing the training results and their performance of all controllers in terms of accuracy and speed, which were represented by the lateral localization accuracy and training time speed.

%With the advantages of DVS for data acquisition, our controllers tends to be fast and robust from illumination conditions. 
%Further, this algorithm is capable of learning to follow different road patterns, even if they are changing within a single scenario. 
%Finally, comparing to the static SNN controller, the proposed algorithm exhibits better performance in terms of the deviation of the robot from the center line.
%
%Building energy-efficient mobile robots using event-based SNNs requires suitable training algorithms.
%In this paper, lane-keeping tasks for Pioneer robot mounted with DVS are tackled by building and training a SNN based controller with the R-STDP learning rule that seeks to bring reinforcement learning capabilities.
%First, with the advantages of DVS for data acquisition, our algorithm tends to be energy efficient and robust from illumination conditions.  
%Further, this algorithm is capable of learning to follow different road patterns, even if they are changing within a single scenario.
%Finally, comparing to the static SNN controller, the proposed algorithm exhibits better performance in terms of the deviation of the robot from the center line.

For future research, the R-STDP controller is intended to be a first step towards more sophisticated algorithms with real reinforcement learning capabilities. 
To date, research has not incorporated reward prediction errors yet, even though this phenomenon was observed in the brain.
Therefore, such networks based on R-STDP should also be implemented using deep architectures in the future.
\section*{Appendix}
\label{sec:appendix}
%%%%%%%%%%%%%%%%%%%%%%%%%%%%%%%%%%%%%%%%%%%%%%%%%%%%%%%%%%%%%%%%%%%%%%%%%%%%%%%%%%%%%%%%%%%%%%%%%%%%%
Simulation parameters for each controller are listed in Tables~\ref{tab param dqn}, \ref{tab param dqn-snn}, \ref{tab:param_rstdp}.
\begin{table}[!tp]
	\caption{DQN parameters}\label{tab param dqn}
	\centering
	\begin{tabular}{lll}\toprule%
		\multirow{7}{*}{DQN}  	& network architecture		& 512 - 200 - 200 - 3\\
		& connections			& fully connected\\ 
		& batch size 			& 32	\\
		& update frequency		& 4	\\
		& soft update			& $\tau=0.001$\\
		& learning rate			& $\alpha = 0.0001$\\
		& buffer size			& 5000 \\
		\toprule
		\multirow{4}{*}{$\epsilon$-greedy policy}	& pre-training steps		& 1000	\\
		& annealing steps		& 49000	\\
		& random probability start	& 1.0	\\
		& random probability end	& 0.1	\\
		\toprule
		\multirow{4}{*}{MDP}		& discount factor	& $\gamma = 0.99$	\\
		& reset distance	& 0.5 m	\\
		& maximum episode steps	& 1000 \\
		& time step length	& $0.5\,s$ \\
		\toprule
		\multirow{2}{*}{Robot}				& motor speed straight	& $v_s = 1.0\, m/s$\\
		& motor speed turn	& $v_t = 0.25\, m/s$\\
		\bottomrule
	\end{tabular}
\end{table}
%%%%%%%%%%%%%%%%%%%%%%%%%%%%%%%%%%%%%%%%%%%%%%%%%%%%%%%%%%%%%%%%%%%%%%%%%%%%%%%%%%%%%%%%%%%%%%%%%%%%%

\begin{table}[!tp]
	\caption{DQN-SNN parameters}\label{tab param dqn-snn}
	\centering
	\begin{tabular}{lll}
		
		\toprule
		\multirow{6}{*}{ANN training}			& network architecture 		& 512 - 200 - 3\\ 
		& connections			& fully connected\\
		& batch size 			& 50	\\
		& training steps		& 10000	\\
		& optimizer			& ADAM\\
		& learning rate			& 0.0001\\
		\toprule
		\multirow{4}{*}{SNN simulation}			& simulation time		& 10 ms	\\
		& max. firing rate		& 1000 Hz\\
		& simulation step length	& 1 ms	\\
		& membrane potential threshold	& 1 mV	\\
		\bottomrule
	\end{tabular}
\end{table}
%%%%%%%%%%%%%%%%%%%%%%%%%%%%%%%%%%%%%%%%%%%%%%%%%%%%%%%%%%%%%%%%%%%%%%%%%%%%%%%%%%%%%%%%%%%%%%%%%%%%%

\begin{table}[!tp]
	\caption{Simulation parameters specification}
	\label{tab:param_rstdp}
	\centering
	\begin{tabular}{lll} \toprule%
		\multirow{5}{*}{Steering model}  & max. speed 			& $v_{max} = 1.5 \, m/s$\\
		& min. speed			& $v_{min} = 1.0 \, m/s$\\
		& turn constant 		& $c_{turn} = 0.5$	\\
		& max spikes during 		& $n_{max} = 15$	\\
		& a simulation step		& \\
		\toprule
		\multirow{3}{*}{Poisson neurons}		& max. firing rate		& 300 Hz\\
		& number of DVS events for	& $n = 15$\\
		& max. firing rate		& \\
		\toprule
		\multirow{2}{*}{SNN simulation}			& simulation time	& 50 ms	\\
		& time resolution	& 0.1 ms\\
		\toprule
		\multirow{12}{*}{LIF neurons}				& NEST model				& iaf\_psc\_alpha\\
		& Resting membrane potential 		& $E_L = -70.0 \, mV$\\
		& Capacity of the membrane 		& $C_m = 250.0 \, pF$\\
		& Membrane time constant		& $\tau_m = 10.0 \, ms$\\
		& Time constant of postsynaptic 	& $\tau_{syn,ex} = 2.0 \, ms$\\
		& excitatory currents			& \\
		& Time constant of postsynaptic  	& $\tau_{syn,in} = 2.0 \, ms$\\
		& inhibitory currents			& \\
		& Duration of refractory period 	& $t_{ref} = 2.0 \, ms$\\
		& Reset membrane potential 		& $V_{reset} = -70.0\,mV$\\
		& Spike threshold			& $V_{th} = -55.0 \, mV$\\
		& Constant input current 		& $I_e = 0.0 \, pA$\\
		\toprule
		\multirow{14}{*}{R-STDP synapse}			& NEST model				& dopamine\_synapse \\
		& Amplitude of weight change		& $A_+ = 1.0$ \\
		& for facilitation			&  \\
		& Amplitude of weight change		& $A_- = 1.0$ \\
		& for depression			& \\
		& STDP time constant	& $\tau_+ = 20.0\, ms$ \\
		& for facilitation	& \\
		& Time constant of	& $\tau_c = 1000.0 \, ms$ \\
		& eligibility trace	&  \\
		& Time constant of 	& $\tau_n = 200.0 \, ms$ \\
		& dopaminergic trace	&  \\
		& Minimal synaptic weight		& $0.0$ \\
		& Maximal synaptic weight		& $3000.0$ \\
		& Initial synaptic weight		& $200.0$ \\
		& Reward constant			& $c_r = 0.01$\\
		\bottomrule
	\end{tabular}
\end{table}

\section*{Acknowledgement}

The research leading to these results has received funding from the Shenzhen Research JCYJ20180507182508857, the European Union Research and Innovation Programme Horizon 2020 (H2020/2014-2020) under the Specic Grant
Agreements No. 720270 (Human Brain Project SGA1), and the Chinese Scholarship Council.

\pagebreak
%\section*{References}

\bibliography{icra2018}
\bibliographystyle{elsarticle-num}

\end{document}